\RequirePackage{amsmath}
\documentclass[runningheads]{llncs}
\usepackage[T1]{fontenc}
\usepackage{graphicx,verbatim}
\usepackage{amsfonts}
\usepackage{makecell}
\usepackage{graphicx}
\usepackage{multirow}
\usepackage{booktabs}
\usepackage{mathrsfs}
\usepackage[export]{adjustbox}
\usepackage{wrapfig}
\usepackage[table]{xcolor}  

\usepackage[backend=biber,
sorting=nyt,
sortcites=true,
url=false,
doi=false,
maxnames=6,
minnames=6,
style=lncs]{biblatex}
\DeclareFieldFormat{titlecase}{\MakeSentenceCase*{#1}}
\newcommand{\figref}[1]{\figurename~\ref{#1}}
\newcommand{\tabref}[1]{\tablename~\ref{#1}}

\DeclareMathOperator{\softmax}{softmax}

\DeclareMathOperator{\TPR}{TPR}

\DeclareMathOperator{\BACC}{BACC}
\DeclareMathOperator{\Fone}{F1}

\DeclareMathOperator{\CE}{ce}
\DeclareMathOperator{\TCE}{tce}
\DeclareMathOperator{\WASS}{w}
\DeclareMathOperator{\WASSCE}{wce}
\DeclareMathOperator{\loss}{\mathcal{L}}

\usepackage[allcolors=black,urlcolor=blue,colorlinks=true]{hyperref}

\begin{document}
\title{Tree-based Semantic Losses: Application to Sparsely-supervised Large Multi-class Hyperspectral Segmentation}
\titlerunning{Tree-based Semantic Losses for Hyperspectral Segmentation}
\author{Junwen Wang\inst{1} \and
Oscar Maccormac\inst{1,2} \and
William Rochford\inst{1,2} \and
Aaron Kujawa\inst{1} \and
Jonathan Shapey\inst{1,2} \and
Tom Vercauteren\inst{1}}
\authorrunning{J. Wang et al.}
%
\institute{
King’s College London, London, UK \\ \email{junwen.wang@kcl.ac.uk}
\and
King's College Hospital, London, UK 
}

\maketitle              
\begin{abstract}
Hyperspectral imaging (HSI) shows great promise for surgical applications, 
offering detailed insights into biological tissue differences beyond what the naked eye can perceive. 
Refined labelling efforts are underway to train vision systems to distinguish large numbers of subtly varying classes. 
However, commonly used learning methods for biomedical segmentation tasks penalise all errors equivalently and thus fail to exploit any inter-class semantics in the label space. In this work, we introduce two tree-based semantic loss functions which take advantage of a hierarchical organisation of the labels.
We further incorporate our losses in a recently proposed approach for training with sparse, background-free annotations.
Extensive experiments demonstrate that our proposed method reaches state-of-the-art performance on a sparsely annotated HSI dataset comprising $107$ classes organised in a clinically-defined semantic tree structure.
Furthermore, our method enables effective detection of out-of-distribution (OOD) pixels without compromising segmentation performance on in-distribution (ID) pixels.
%
\end{abstract}

\section{Introduction}
Hyperspectral imaging (HSI) 
is an emerging optical imaging technique that collects and processes spectral data across a range of wavelengths \cite{shapeyIntraoperativeMultispectralHyperspectral2019}. By splitting light into multiple narrow bands beyond the capability of human vision, HSI captures additional details that are imperceptible to the naked eye. This technique can provide diagnostic information about tissue properties, enabling objective characterisation of tissues without the use of external contrast agents.
Surgical image segmentation is an important application of HSI. The use of deep learning for biomedical segmentation with spectral imaging data is increasing \cite{khanTrendsDeepLearning2021}, alongside the growing availability of medical HSI datasets \cite{studier-fischerHeiPorSPECTRALHeidelbergPorcine2023,hyttinenOralDentalSpectral2020,fabeloHELICoiDProjectNew2016,carstensDresdenSurgicalAnatomy2023}. 

Developing an effective segmentation model for HSI data remains challenging. One reason is that most existing HSI datasets are sparsely annotated in order to reduce the substantial workload~\cite{studier-fischerHeiPorSPECTRALHeidelbergPorcine2023,hyttinenOralDentalSpectral2020,carstensDresdenSurgicalAnatomy2023}. 
In the sparse annotation setting, only a few representative areas of foreground objects are labelled. Examples of this labelling strategy are illustrated in the second column of \figref{fig:qualitative_results_all_classes}.
In a recent study, Wang et al.~\cite{wang2024oodsegoutofdistributiondetectionimage} introduced a medical image segmentation framework that enables pixel-level out-of-distribution (OOD) detection using sparsely annotated data containing only positive classes. Their method uses these sparse annotations to learn robust feature representations, which subsequently facilitate the reliable detection of OOD pixels via conventional OOD detection techniques originally developed for classification tasks. This approach achieves state-of-the-art OOD detection performance while preserving classification accuracy for in-distribution (ID) classes.

Another challenge relates to the granularity at which the data is annotated and that at which segmentation models should operate. 
Recent works have examined the impact of training models with 
various levels of labelling granularity -- such as pixels, patches, and entire images \cite{seidlitzRobustDeepLearningbased2022,garciaperazaherreraHyperspectralImageSegmentation2023}. Other studies have provided comparative analyses of different pixel-level algorithms for brain tissue differentiation \cite{martin-perezMachineLearningPerformance2024}.
Underpinning these questions is the drive for a holistic and refined understanding of the surgical scenes.
Annotation efforts are ongoing to provide training data across large number of potentially subtly varying classes.
Combined with sparse annotations processes, many of these classes may only have small amounts of training samples.
It is thus important to take advantage of the semantics of the labels and realise that some types of errors are more acceptable than others.
However, no previous work in the field of surgical imaging has leveraged the structure of the label space as a source of information. 
The importance of label semantics is getting recognised in the general field of computer vision~\cite{frognerLearningWassersteinLoss2015a,leTreeSlicedVariantsWasserstein2019a,bertinettoMakingBetterMistakes2020}.
In the task of brain tumour segmentation, Fidon et al.~\cite{fidonGeneralisedWassersteinDice2018} proposed a variant of the Dice score for multi-class segmentation based on the Wasserstein distance in the probabilistic label space.
However, this method was only demonstrated in the context of a small number of labels from the BraTS challenge~\cite{menzeMultimodalBrainTumor2015}, and it does not generalise to sparse, background-free annotations.


In this work, we propose leveraging prior knowledge of label structures through two tree-based semantic losses for sparsely supervised segmentation.
By incorporating a comprehensive label hierarchy curated by domain experts, the model achieves superior performance compared with the standard cross-entropy baseline.
Our proposed tree-based hierarchical weighting scheme establishes a new state-of-the-art multi-class hyperspectral image segmentation method.
Furthermore, we integrate these losses into the OOD detection framework of \cite{wang2024oodsegoutofdistributiondetectionimage} to enable OOD detection without compromising performance on ID data.

\section{Material and methods}
\subsubsection{Surgical imaging dataset}
\begin{figure}[bt]
    \centering
    \includegraphics[width=\textwidth]{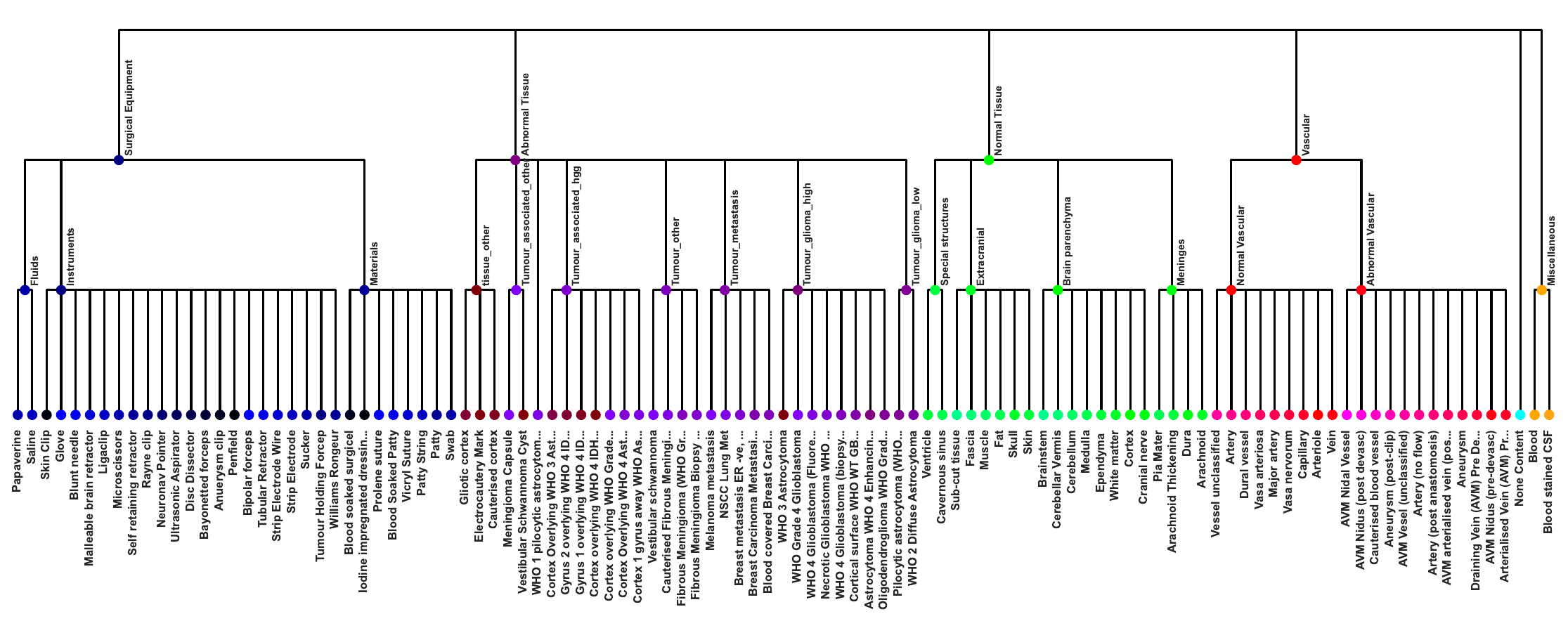}
    \caption{Full tree-based label hierarchy of the surgical HSI dataset. From top to bottom, the hierarchy progresses from coarse object categories to specific classes. The colour coding matches the ground-truth mask at each level.}
    \label{fig:hierarchy}
\end{figure}
The data was obtained from patients undergoing microscopic cranial neurosurgery as part of an ethically approved single-centre, prospective clinical observational investigation employing a prototype hyperspectral imaging system
(NeuroHSI study: REC reference 22/LO/0046, ClinicalTrials.gov ID NCT05294185). 
Informed consent was obtained from all participants. The primary objective was to evaluate the intraoperative utility of a \(4 \times 4\), 16-band visible-range snapshot mosaic camera (IMEC CMV2K-SSM4X4-VIS).

The dataset comprises 22,829 annotated frames derived from 45 distinct surgical cases, encompassing both neuro-oncological and neurovascular pathologies. Multiple videos were acquired for each case, with each recording representing a specific surgical phase intended to capture relevant intraoperative details. Individual phases included varying visual perspectives to account for changes in camera positioning during video acquisition. Training snapshot data are first processed with a demosaicking pipeline \cite{liDeepLearningApproach2022, Li_2024_BMVC}, resulting in frames comprising $16 \times 1088 \times 2048$ pixels enabling reconstruction of synthetic sRGB visualisation from the HSI data.

The sparse, background-free annotations encompass 107 subclasses organised into a hierarchical structure defined by neurosurgeons. \figref{fig:hierarchy} presents the complete label hierarchy, including four primary categories at the top level of the tree: surgical equipment, abnormal tissue, normal tissue, and vascular structures with corresponding colour references displayed at each node in the label hierarchy.
A subset of frames from each video was manually annotated by an experienced neurosurgeon, and where feasible, these annotations were then propagated across subsequent frames algorithmically using the registration-based propagation feature of the ImFusion Labels software. Each propagated annotation was verified and corrected (when needed) by a neurosurgeon prior to final submission.

In addition to human-labelled categories, an additional label was generated by a content area estimation algorithm \cite{buddRapidRobustEndoscopic2023}. This algorithm provides a robust estimation of non-informative areas near the edge of the image. By treating these areas as part of our label hierarchy, the model can more effectively discriminate content regions. \figref{fig:qualitative_results_all_classes} presents example sRGB image-annotation overlays.


\subsubsection{Wasserstein distance in label space}
Let $\mathbf{L}$ be the label space (e.g. as in \figref{fig:hierarchy}) with $C$ leaf nodes, where $\mathbf{L} = \{1,2,...,C \}$. 
Let $p,q \in P(\mathbf{L})$ be probability vectors on $\mathbf{L}$.
The Wasserstein distance between $p$ and $q$ is the minimal cost to transform $p$ into $q$ given the ground distances $M_{l,l'}\in\mathbb{R}^+$ between any two labels $l$ and $l'$. 
The ground distance is represented as a matrix $M$ and the associated Wasserstein distance $W^M(p,q)$ is defined through an optimal transport problem:
\begin{equation}\label{eq:wassdist}
	\begin{split}
		W^M(p,q) &= \min_{T_{l,l'}}\sum_{l,l'\in\mathbf{L}} T_{l,l'}M_{l,l'} \\
		\textrm{subject to } \forall l\in\mathbf{L},\sum_{l'\in\mathbf{L}} T_{l,l'} &= p_l
        \quad \textrm{and} \quad 
            \forall l'\in\mathbf{L},\sum_{l\in\mathbf{L}} T_{l,l'} = q_{l'}
	\end{split}
\end{equation}
By leveraging the distance matrix $M$ on $\mathbf{L}$, the Wasserstein distance yields a semantically-meaningful way of comparing two label probability vectors. 
Given a tree structure as in \figref{fig:hierarchy} with weights associated to the edges, a semantic ground distance can be induced by the path lengths between the leaf nodes.
If $q=g$ is a \emph{crisp} ground truth, a closed-form expression of \eqref{eq:wassdist} is given in \cite{fidonGeneralisedWassersteinDice2018}:
\begin{equation}\label{eq:wassdistcrisp}
	W^{M} (p, g) = \sum_{l,l'\in\mathbf{L}} M_{l,l'}p_l g_{l'} = p^T M g
\end{equation}

Supervised training of segmentation models has been optimised for
pixel-level cross-entropy (CE) which does not account for any label semantics:
\begin{equation}\label{eq:ce}
    CE (p, g) =  - \sum_{l} g_l\log(p_l)
\end{equation}
Several works have shown benefits in combining CE with task-specific losses \cite{maLossOdysseyMedical2021}.
We thus propose a simple combination of cross-entropy and Wasserstein distance which we refer to as \textit{Wasserstein+CE} loss:
\begin{equation}\label{eq:wassce}
	\loss_{\WASSCE}^{M} (p, g) = \alpha CE + \beta W^{M}
\end{equation}
where we set $\alpha = \beta = 0.5$ across all experiments. 
In the case of $\alpha = 0$ and $\beta = 1$, equation \eqref{eq:wassce} reduces to the vanilla Wasserstein distance $\loss_w^M$.
In the case of $\alpha = 1$ and $\beta = 0$, equation \eqref{eq:wassce} reduces to standard CE loss $\loss_{ce}$.

\subsubsection{Tree-based semantic cross-entropy loss}
We propose a novel semantic loss function which takes advantage of the tree structure, computes the (aggregated) probabilities across the nodes (not just the leaf nodes), and evaluates an extended CE loss on all node probabilities.
Let the label tree $\mathcal{T}$ be composed of $K$ levels with $0$ corresponding to the deepest level (leaf nodes).
Let $A$ be the adjacency matrix associated with $\mathcal{T}$. 
Let $\tilde{p}$ be a zero-padding of $p$ to initially associate non-leaf nodes with a zero mass, and $p^{\dagger}$ be the vector collecting all the probabilities:
\begin{equation}\label{eq:collected_prob}
	p^{\dagger} = (\sum_{k \geq 0} A^k ) \tilde{p}
\end{equation}
where $A^k = 0$ for $k > K$. We express the extended weighted CE as:
\begin{equation}\label{eq:cetree}
	CE^{\mathcal{T}}(p,g) = - \sum_v w_v g^{\dagger}_v\log(p^{\dagger}_v)
\end{equation}
where $w_v$ is the weight of the edge associated with $v$ as a child node. 
We note that if $w_v=1$ for all leaf nodes and $w_v=0$ otherwise, equation \eqref{eq:cetree} reduces to \eqref{eq:ce}. 
We refer to our semantically-informed variant of the CE loss as \textit{tree-based semantic cross-entropy} loss $\loss^{M}_{\TCE}$.


\subsubsection{Out-of-distribution segmentation}
We build on the recent work \cite{wang2024oodsegoutofdistributiondetectionimage} to segment OOD samples from sparse multi-class positive-only annotations. Given a 2D image $x$, each annotated spatial location $(i,j)$ from $x$ has a corresponding annotation $y_{ij}$, where $y_{ij} \in\{c\}=\{1,2,\dots,C\}$ and $C$ is the number of classes that marked as ID. 
$c=0$ is retained to denote OOD pixels.
Our proposed framework employs a confidence threshold $\tau$ to detect OOD samples at the pixel level:
\begin{equation}
	\hat{y}_{ij} =
	\begin{cases}
		\operatorname*{arg\,max}_c \boldsymbol{S}_{ij}^c, & \text{if } \max_c \boldsymbol{S}_{ij}^c > \tau, \\
		0, & \text{otherwise},
	\end{cases}
	\label{eqa:ood_decision}
\end{equation}
where $\boldsymbol{S}_{ij}^c$ is a scoring function that captures the probability of pixel $(i,j)$ belonging to the ID class $c$, while acknowledging the possibility of it being OOD in the input image. 
$\boldsymbol{S}_{ij}^c$ can be selected from various OOD detection methods used in image classification tasks \cite{hendrycksBaselineDetectingMisclassified2017,liangEnhancingReliabilityOutofdistribution2018,leeSimpleUnifiedFramework2018,hsuGeneralizedODINDetecting2020}. 
For simplicity, we let $\boldsymbol{S}$ be the network output probabilities aggregated at level $k$ i.e. $\boldsymbol{S}_{ij} = p^\dagger_k$. The collected probabilities $p^\dagger$ are computed using \eqref{eq:collected_prob} based on the leaf node's softmax probability $p = \softmax\bigl(f(x)\bigr)_{ij}$, where $f(x)$ is the segmentation network with input $x$, trained with positive-only sparse annotation.

\section{Experimental results}
\subsubsection{Implementation details}
For baseline (cross-entropy) training, we adopted a similar training pipeline as described in \cite{wang2024oodsegoutofdistributiondetectionimage}. Specifically, we used a U-Net architecture with an EfficientNet-b5 encoder \cite{tanEfficientNetRethinkingModel2019}, pre-trained\footnote{\url{https://github.com/qubvel/segmentation_models.pytorch}} on ImageNet \cite{dengImageNetLargescaleHierarchical2009} for all experiments. 
We conducted model training using the softmax output and a one-hot encoded sparse annotation mask for pixels marked as ID. We employed the Adam optimiser \cite{kingmaAdamMethodStochastic2017} with $\beta_1 = 0.9$ and $\beta_2 = 0.999$, together with an exponential learning rate scheme ($\gamma = 0.999$). We set the initial learning rate to $0.0001$, used a mini-batch size of $5$, and trained for a total of $50$ epochs. For a fair comparison, we applied the same hyperparameters across all methods. 
For data augmentation, we adopted a similar setup to that reported in \cite{wang2024oodsegoutofdistributiondetectionimage}: random rotation (rotation angle limit: $45^\circ$), random flipping, random scaling (scaling factor limit: $0.1$), and random shifting (shift factor limit: $0.0625$). All transformations were applied with a probability of $0.5$. In addition, we apply $\ell^1$-normalisation at each spatial location $ij$ to account for the non-uniform illumination of the tissue surface. This is routinely applied in HSI because of the dependency of the signal on the distance between the camera and the tissue \cite{bahlSyntheticWhiteBalancing2023,studier-fischerHeiPorSPECTRALHeidelbergPorcine2023}.
For Wasserstein-based training, we used the same hyperparameters as those used in the baseline approach to ensure a fair comparison.
Experiments were run on an NVidia DGX cluster with V100 (32GB) and A100 (40GB) GPUs.

\subsubsection{Choice of tree weights / ground distances}
Because edge weights exist at different hierarchical levels, resulting in distinct matrices $M$ that affect overall performance,
we investigated the impact of various edge weight strategies for $M$.
Specifically, $M_{t}$ (respectively  $M_{\ell}$) represent a configuration in which the edge weight is set to $1$ only at the top level (respectively leaf node) with $0$ elsewhere.
$M_{e}$ and $M_{h}$ denote configurations in which edge weights are assigned at all levels, with different ratios. In particular, $M_e$ sets all edge weights to $1$, while $M_h$ assigns weights in a 100--10--1 ratio to the top, middle, and bottom level edges, respectively. The latter encodes expert intuition that errors high in the hierarchy are clinically more severe. 
For clarity, we note that the proposed $\loss_{\TCE}$ with $M_{\ell}$ is equivalent to standard CE loss $\loss_{\CE}$.
Beyond handcrafted choice, a more systematic approach would tune the edge weights automatically, such as with hyperparameter optimisation to search over candidate configurations \cite{liaw2018tune}. 

\subsubsection{Cross-validation results}
\begin{table}[bt]
    \centering
    \setlength\tabcolsep{4pt}
    \caption{Cross-validation results on top-level classes. 
    For each method and metric, performance is reported at thresholds $\tau_0=0$ and $\tau_m$. 
    $\tau_m$ is chosen from the optimal threshold with highest performance in the validation set. 
    The best performance among all losses is highlighted in bold. 
    Rows shaded in grey represent the baseline results, which are equivalent to the standard CE or Wasserstein+CE training on leaf node classes only (i.e. without class semantics). 
    The asterisk $``*"$ indicates a strong baseline result which is equivalent to standard CE training on top-level nodes only.}
    \begin{tabular}{c c cc cc cc}
    \toprule
    \multirow{2}{*}{\textbf{Loss}} & 
     &
    \multicolumn{2}{c}{$\boldsymbol{\TPR}\uparrow$} &
    \multicolumn{2}{c}{$\boldsymbol{\BACC}\uparrow$} &
    \multicolumn{2}{c}{$\boldsymbol{\Fone}\uparrow$} \\
    \cmidrule(lr){3-4}
    \cmidrule(lr){5-6}
    \cmidrule(lr){7-8}
    \multicolumn{2}{c}{} & $\boldsymbol{\tau_0=0}$ & $\boldsymbol{\tau_m}$
                         & $\boldsymbol{\tau_0=0}$ & $\boldsymbol{\tau_m}$
                         & $\boldsymbol{\tau_0=0}$ & $\boldsymbol{\tau_m}$
                         \\
    \midrule
    $\loss_{\WASS}$ & $M_{t}$ & $0.51_{\pm0.03}$ & $0.51_{\pm0.03}$ & $0.74_{\pm0.01}$ & $0.74_{\pm0.01}$ & $0.47_{\pm0.05}$ & $0.47_{\pm0.05}$ \\
    \midrule
    \rowcolor{gray!30} \multirow{4}{*}{$\loss_{\TCE}$} & $M_{\ell}$ & $0.61_{\pm0.04}$ & $0.65_{\pm0.05}$ & $0.79_{\pm0.02}$ & $0.82_{\pm0.02}$ & $0.60_{\pm0.02}$ & $0.65_{\pm0.03}$ \\
     & $M_{e}$ & $0.64_{\pm0.04}$ & $0.76_{\pm0.03}$ & $0.80_{\pm0.02}$ & $0.88_{\pm0.01}$ & $0.62_{\pm0.05}$ & $0.74_{\pm0.05}$ \\
     &$^*M_{t}$ & $0.65_{\pm0.05}$ & $0.73_{\pm0.12}$ & $0.81_{\pm0.02}$ & $0.86_{\pm0.06}$ & $0.63_{\pm0.04}$ & $0.72_{\pm0.10}$ \\
     & $M_{h}$ & $0.65_{\pm0.03}$ & $0.75_{\pm0.03}$ & $0.81_{\pm0.02}$ & $0.87_{\pm0.02}$ & $0.64_{\pm0.04}$ & $0.76_{\pm0.05}$ \\
    \midrule
    \rowcolor{gray!30} \multirow{4}{*}{$\loss_{\WASSCE}$} & $M_{\ell}$ & $0.61_{\pm0.06}$ & $0.70_{\pm0.05}$ & $0.79_{\pm0.03}$ & $0.85_{\pm0.03}$ & $0.62_{\pm0.04}$ & $0.72_{\pm0.04}$ \\
     & $M_{e}$ & $0.65_{\pm0.04}$ & $0.72_{\pm0.09}$ & $0.81_{\pm0.02}$ & $0.85_{\pm0.04}$ & $0.64_{\pm0.01}$ & $0.73_{\pm0.07}$ \\
     & $M_{t}$ & $0.66_{\pm0.04}$ & $0.77_{\pm0.07}$ & $0.81_{\pm0.02}$ & $0.88_{\pm0.04}$ & $0.63_{\pm0.03}$ & $0.78_{\pm0.09}$ \\
     & $M_{h}$ & $\boldsymbol{0.68}_{\pm0.02}$ & $\boldsymbol{0.80}_{\pm0.03}$ & $\boldsymbol{0.83}_{\pm0.01}$ & $\boldsymbol{0.90}_{\pm0.02}$ & $\boldsymbol{0.66}_{\pm0.02}$ & $\boldsymbol{0.82}_{\pm0.06}$ \\
    \bottomrule
    \end{tabular}
    \label{tab:cross_validation}
\end{table}

\tabref{tab:cross_validation} presents the cross-validation results on top-level classes by selecting the output probability at the top level only (i.e., $p^\dagger_K$). We report the results with different confidence thresholds $\tau$. 
Where $\tau_0=0$ represents no OOD detection and $\tau_m$ represents the threshold which maximises ID score. 
Potentially $\tau_m$ can be also computed by incorporating held-out OOD classes in the two-level cross-validation \cite{wang2024oodsegoutofdistributiondetectionimage}. 
However, this is not presented here due to time constraints in running cross-validation with held-out OOD class.
For all loss functions we evaluate the \textit{True Positive Rate (TPR)}, \textit{Balanced Accuracy (BACC)}, and \textit{F1 scores}. Results are reported by averaging across classes under one-vs-rest strategy, where the positives are pixels of such class, and the negative are the pixels of all the other classes. 
For statistical evaluation, we conducted paired-sample t-test on the test image scores. To limit the number of paired comparisons, we focus on F1-score and comparison between the final tree-based results ($M_h$ configuration for $\loss_{tce}$ or $\loss_{wce}$) and their corresponding baselines ($M_{\ell}$ configuration). Both comparisons were found statistically significant ($p < 0.0001$).
For model performance on leaf node classes, we report F1 scores at $\tau_m$ for both losses. For $\loss_{\WASSCE}$, the mean of F1 scores at $\tau_m$ based on $M_{\ell}$ and $M_h$ are $0.069$ and $0.073$, respectively. For $\loss_{\TCE}$, the results are $0.068$ and $0.037$, respectively.

While both loss outperforms the baseline, our results shows that $M_{t}$ performs similarly to $M_{e}$, suggesting that top-level edge weights have the greatest impact on performance when evaluating accumulated probabilities on corresponding nodes. 
For $\loss_{\WASSCE}$, employing $M_h$ yields the best performance, even surpassing the strong baseline $M_t$, demonstrating that an optimal $M$ can 
achieve state-of-the-art results on both top-level and leaf nodes.
For OOD segmentation, our findings exhibit trends similar to those reported in previous work on three medical image datasets \cite{wang2024oodsegoutofdistributiondetectionimage}. By removing pixels considered outliers, all methods gain further improvements.

\subsubsection{Error analysis}
\begin{figure}[tb]
    \centering
    \includegraphics[width=\linewidth]{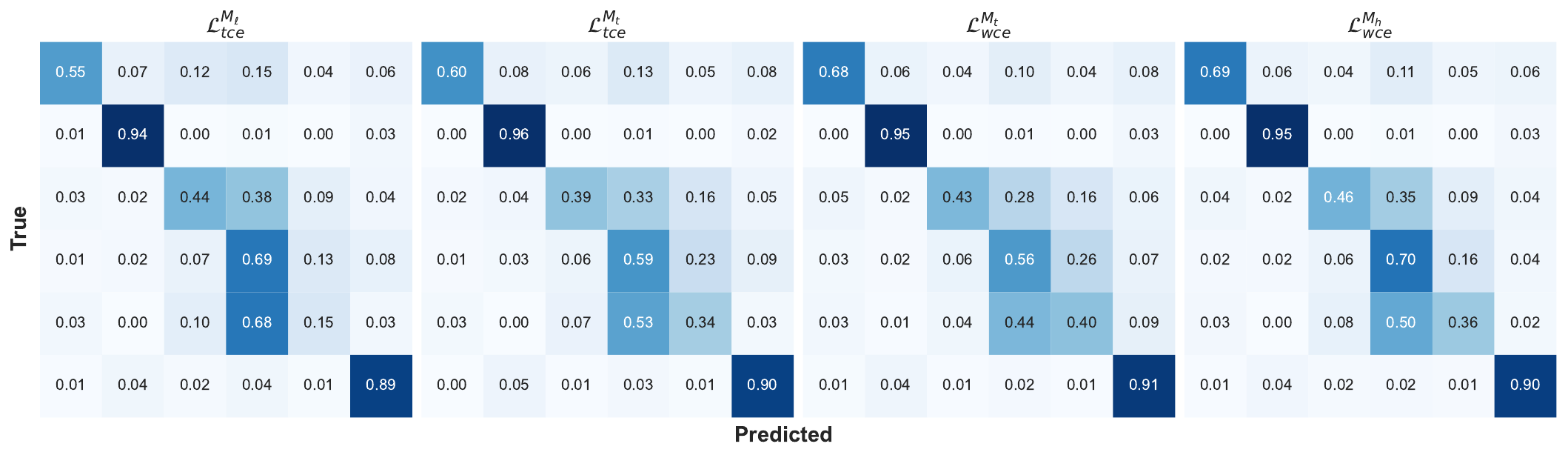}
    \caption{Confusion matrices on top-level nodes. 
    Results are averaged across all cross-validation folds. Class names from top-left to bottom-right: Other, Out-of-focus Area, Vascular, Normal Tissue, Abnormal Tissue and Surgical Equipment.}
    \label{fig:confusion_matrix}
\end{figure}

Evaluation relying solely on overall segmentation performance is insufficient to capture how effectively a model infers relationships among different tissue types. 
It neglects the possibility that the model may choose semantically coherent labels, even if these do not precisely align with the ground truth. 
To explore these aspects, we plot a multi-class confusion matrix for the top-level nodes $p^\dagger_K$, as shown in \figref{fig:confusion_matrix}. 
Because class distributions vary across folds, we average the results of each cross-validation fold to ensure a fair comparison.

Under the cross-entropy baseline, the model struggles to distinguish normal from abnormal tissues, which constitute a semantically important distinction. 
By contrast, both tree-based semantic losses (excluding ${M_{\ell}}$ cases) exhibit a more meaningful confusion pattern between normal and abnormal tissues. 
These findings suggest that the proposed method successfully exploits hierarchical relationships within the label space.

\subsubsection{Qualitative results}
\begin{figure}[tb]
\centering
\setlength\tabcolsep{0pt} 
\begin{tabular}{c c c c c c c c}
    & \textbf{sRGB} 
    & \makecell[c]{\textbf{\tiny Sparsely}\\[-5pt]\textbf{\tiny annotated}\\[-5pt]\textbf{\tiny ground truth}}
    & \textbf{$\loss_{\TCE}^{M_{\ell}}(\tau_{0})$}
    & \textbf{$\loss_{\TCE}^{M_{\ell}}$}
    & \textbf{$\loss_{\TCE}^{M_{t}}$}
    & \textbf{$\loss_{\WASSCE}^{M_{t}}$}
    & \textbf{$\loss_{\WASSCE}^{M_{h}}$}\\
    
    & \includegraphics[width=.14\linewidth,valign=m]{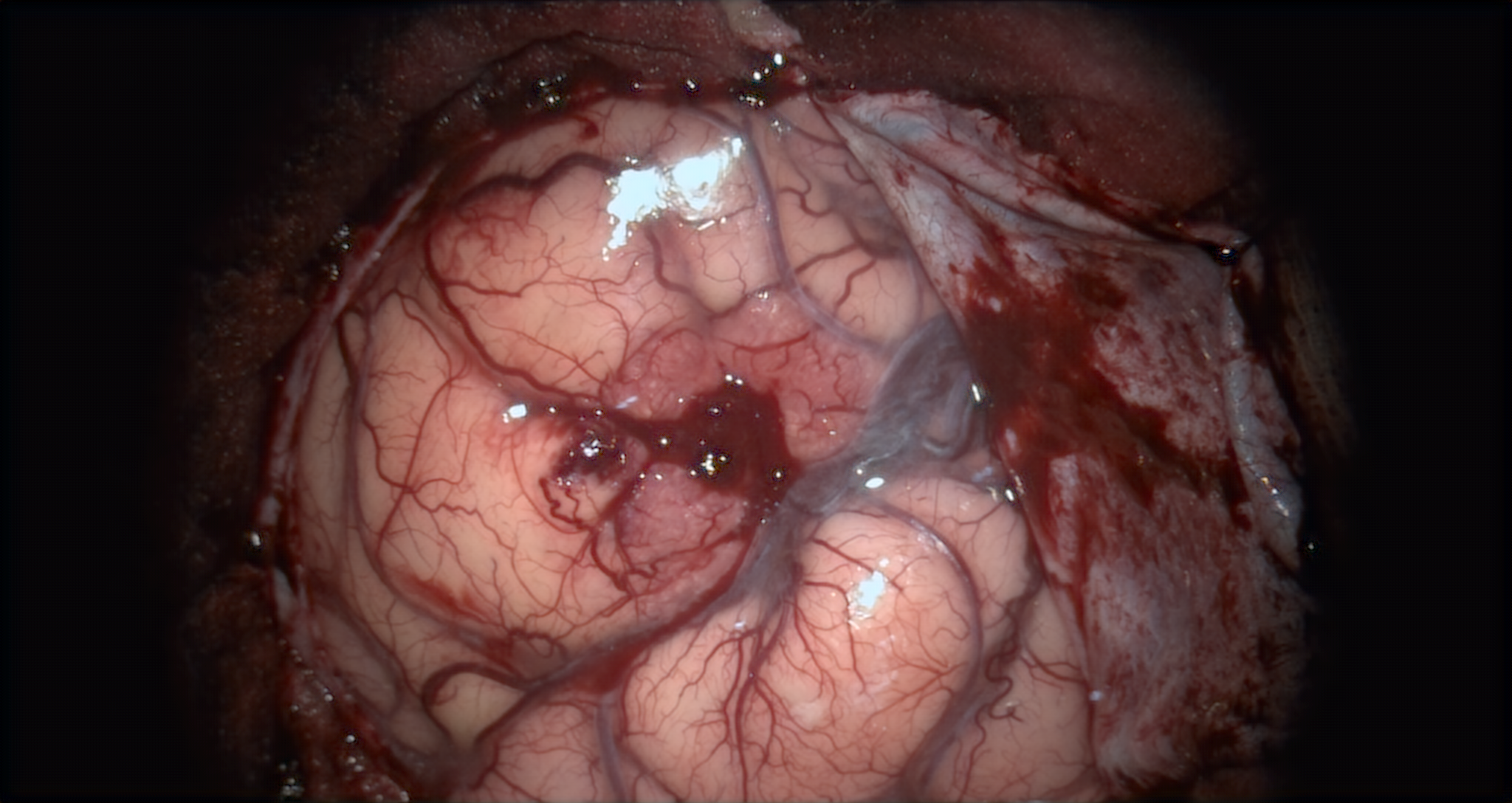}
    & \includegraphics[width=.14\linewidth,valign=m]{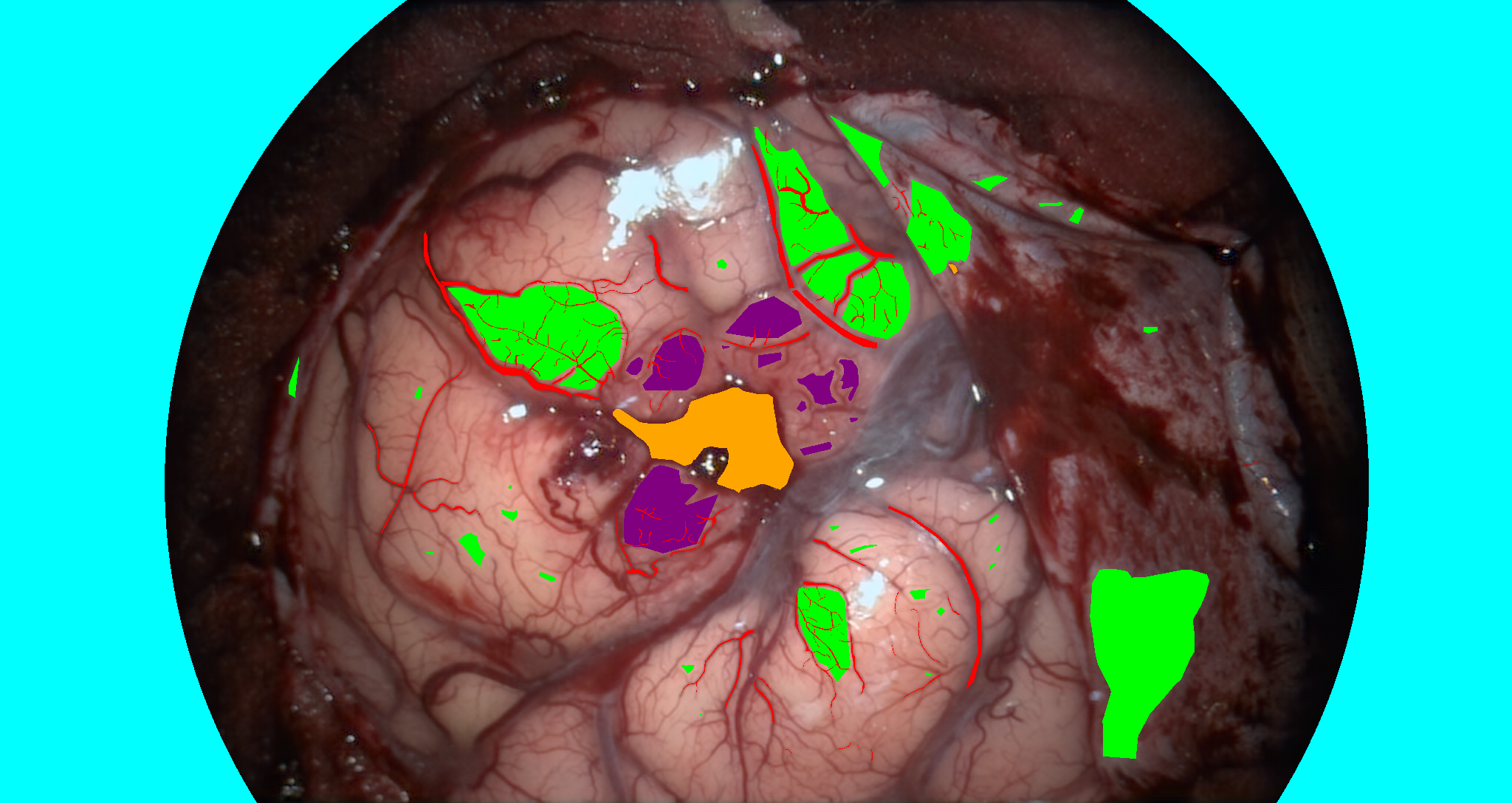}
    & \includegraphics[width=.14\linewidth,valign=m]{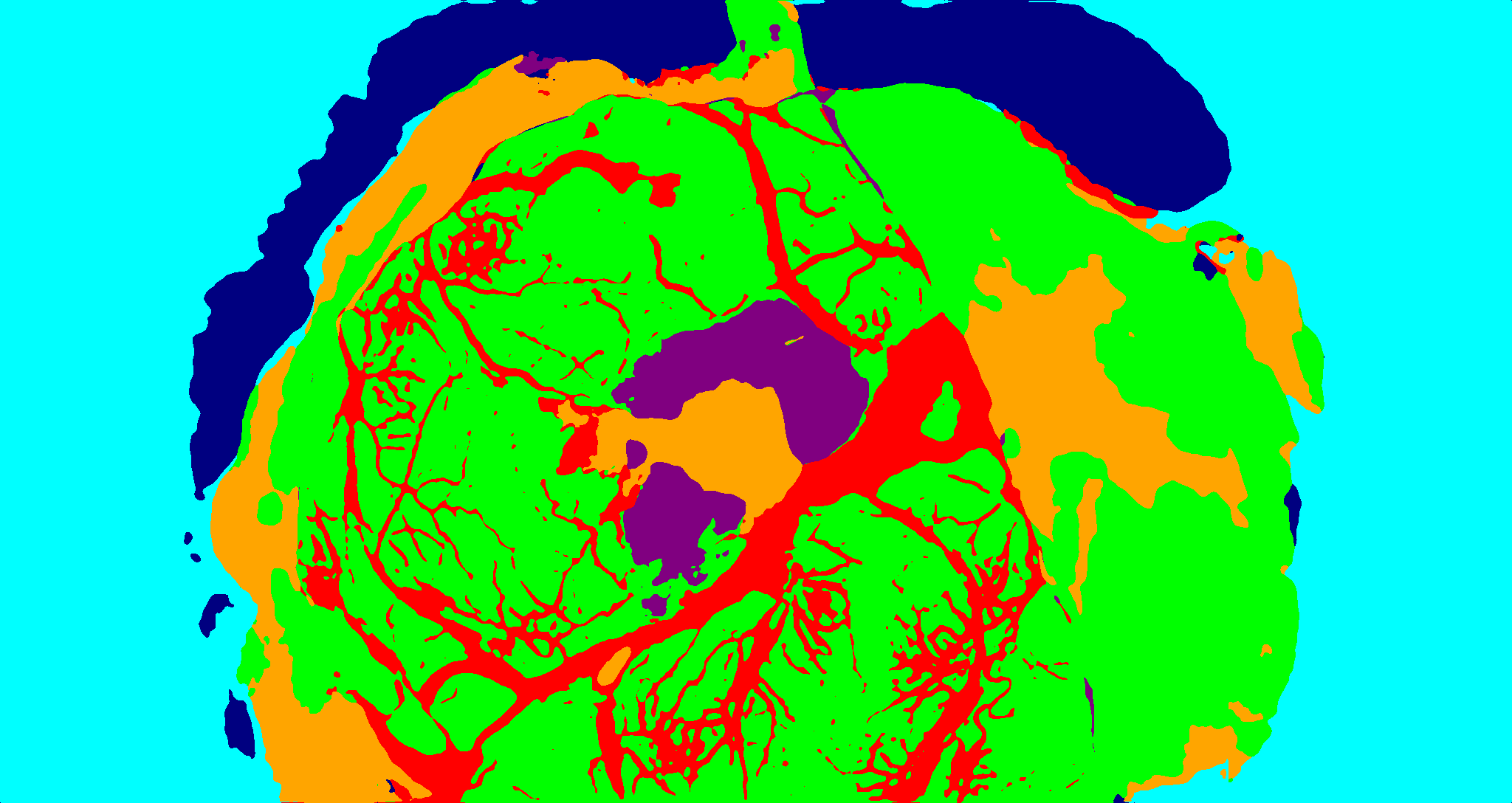}
    & \includegraphics[width=.14\linewidth,valign=m]{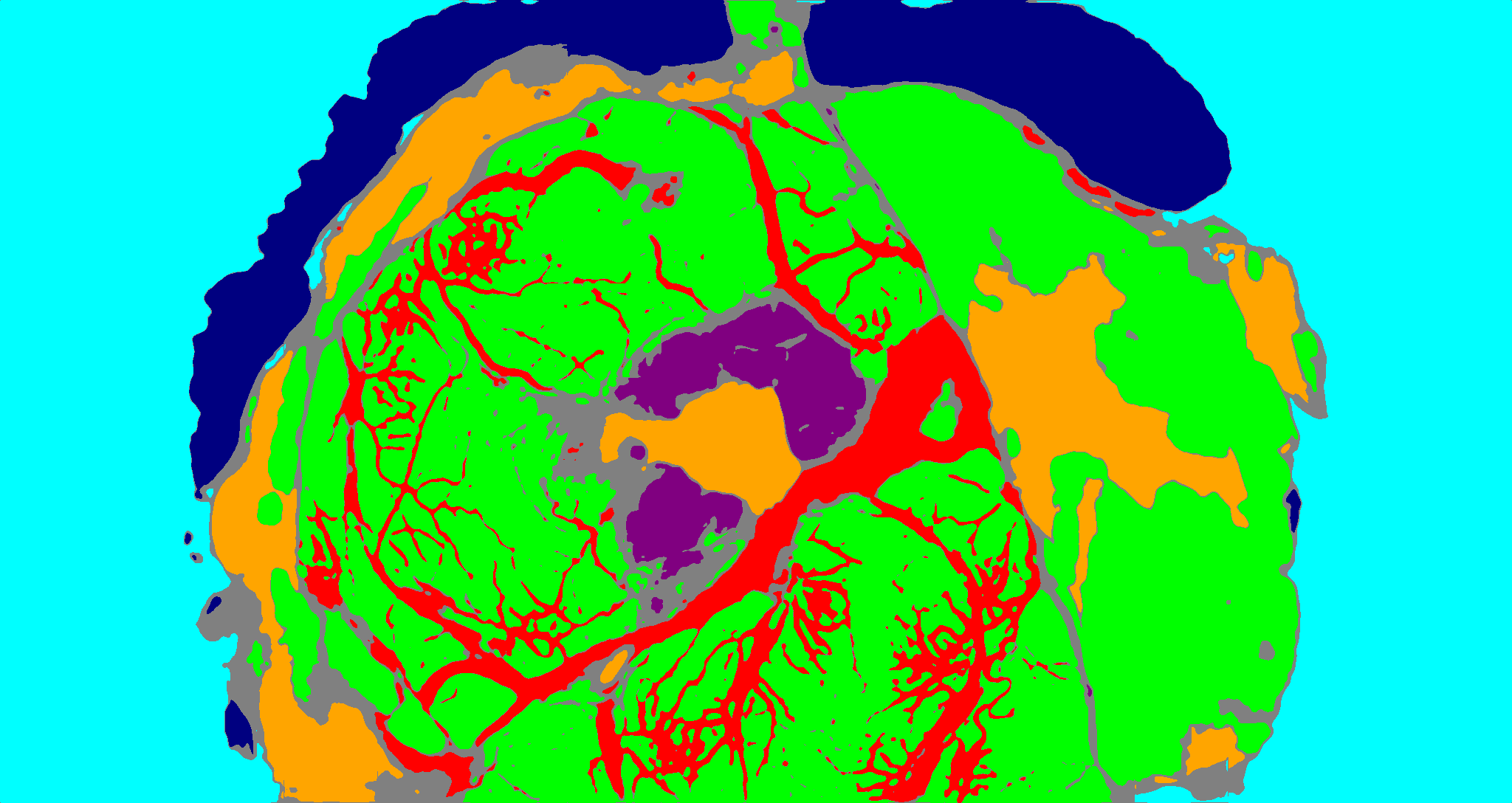}
    & \includegraphics[width=.14\linewidth,valign=m]{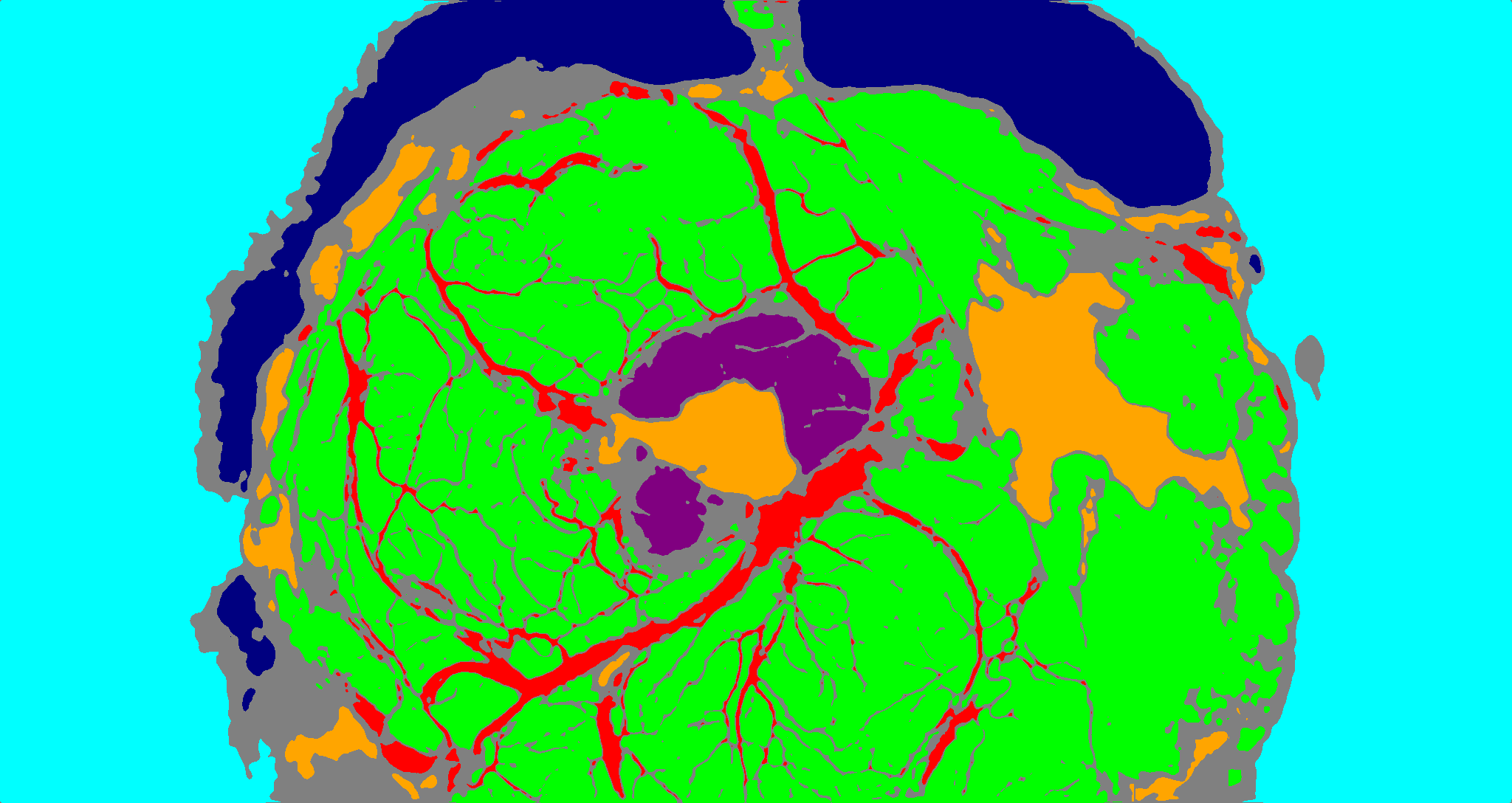}
    & \includegraphics[width=.14\linewidth,valign=m]{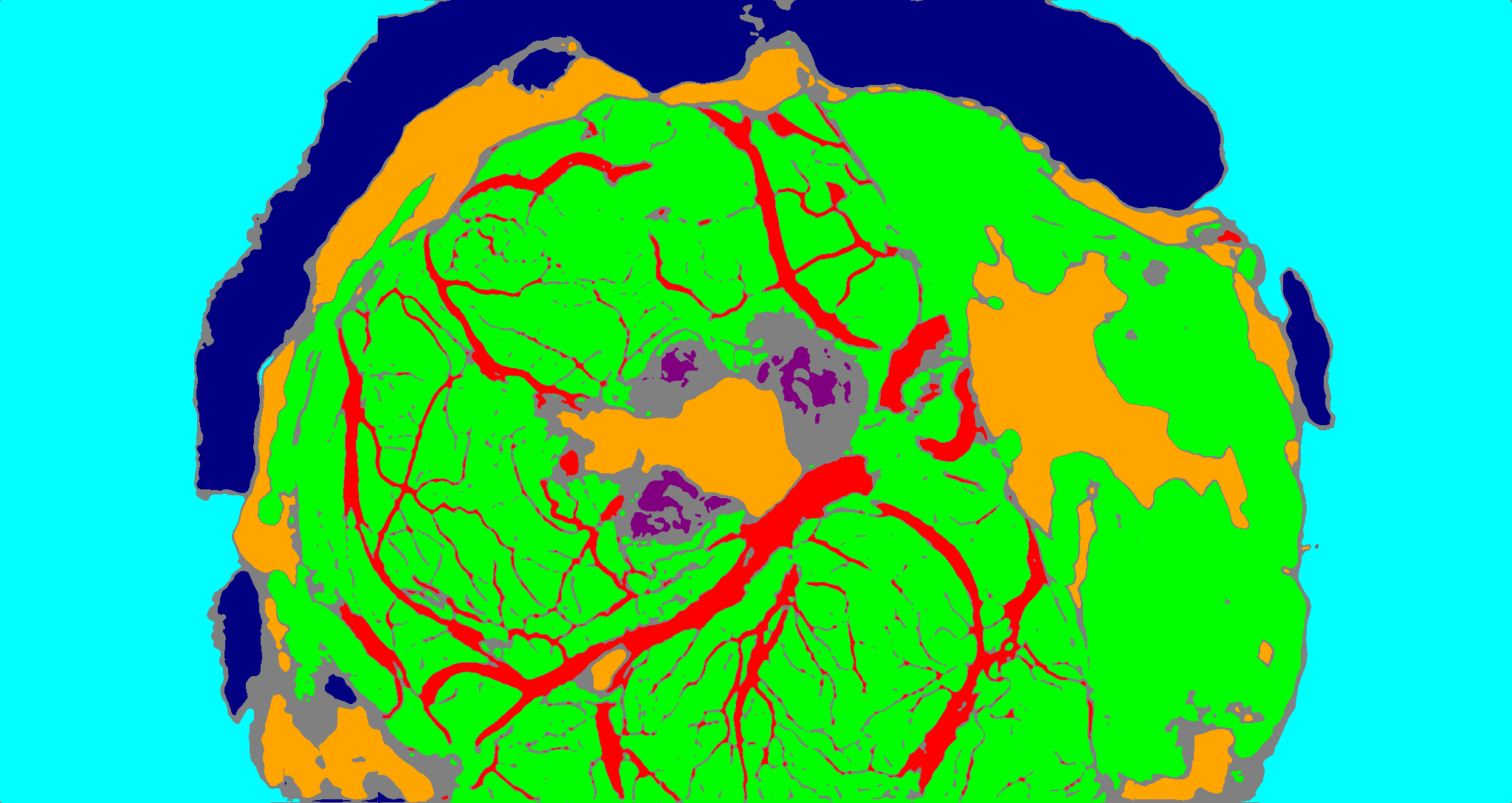}
    & \includegraphics[width=.14\linewidth,valign=m]{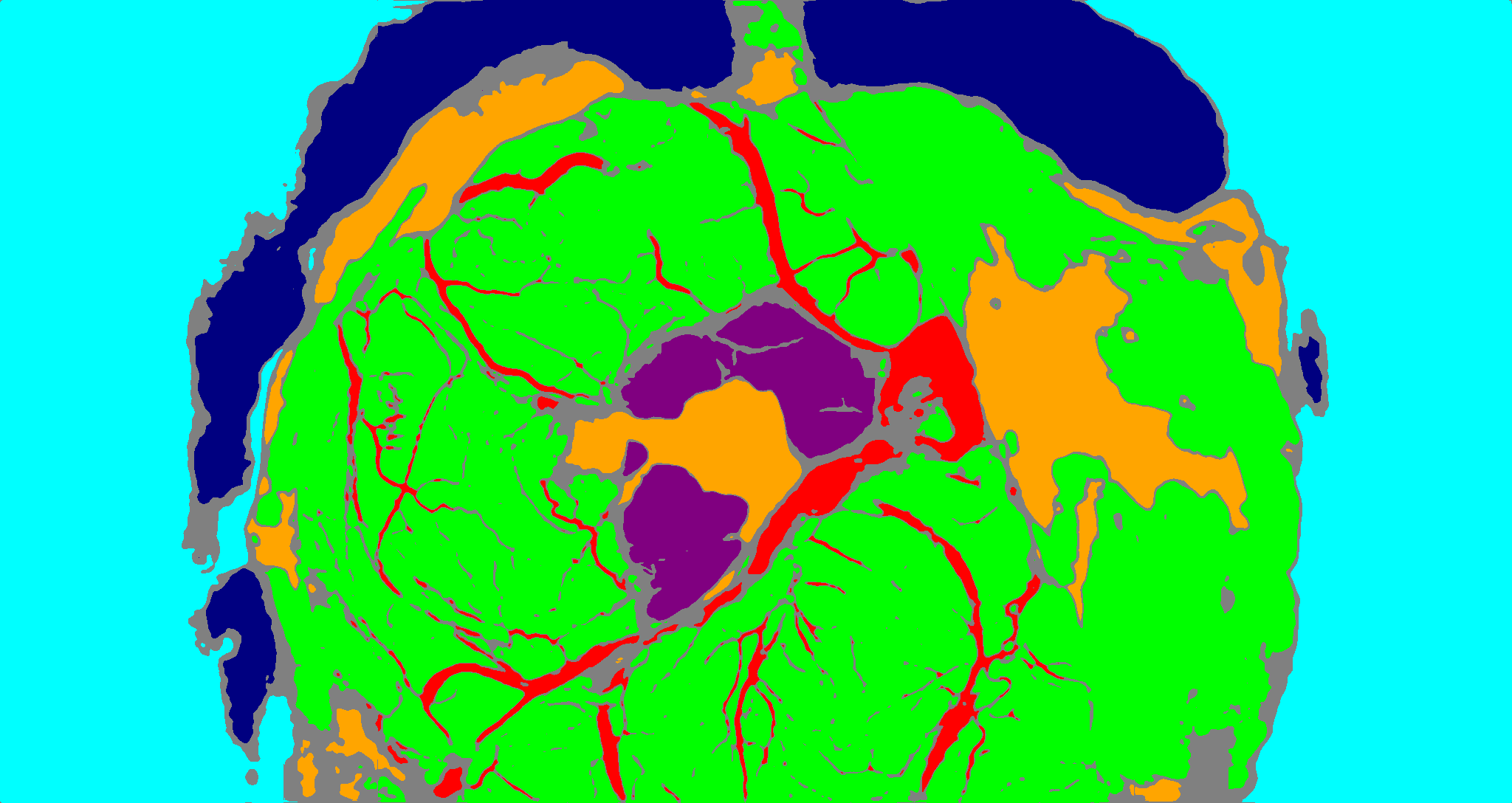}
    \\

    & \includegraphics[width=.14\linewidth,valign=m]{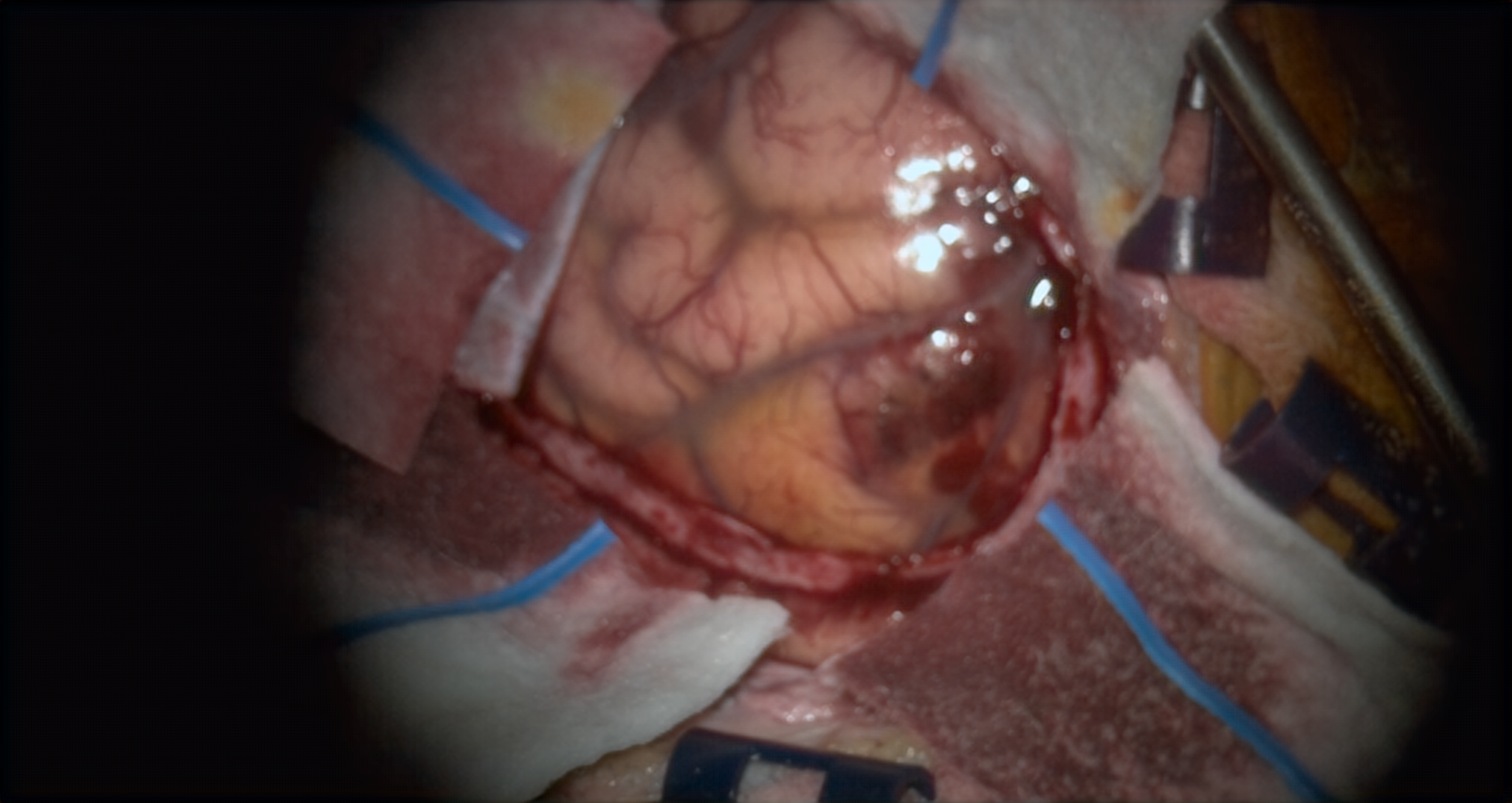}
    & \includegraphics[width=.14\linewidth,valign=m]{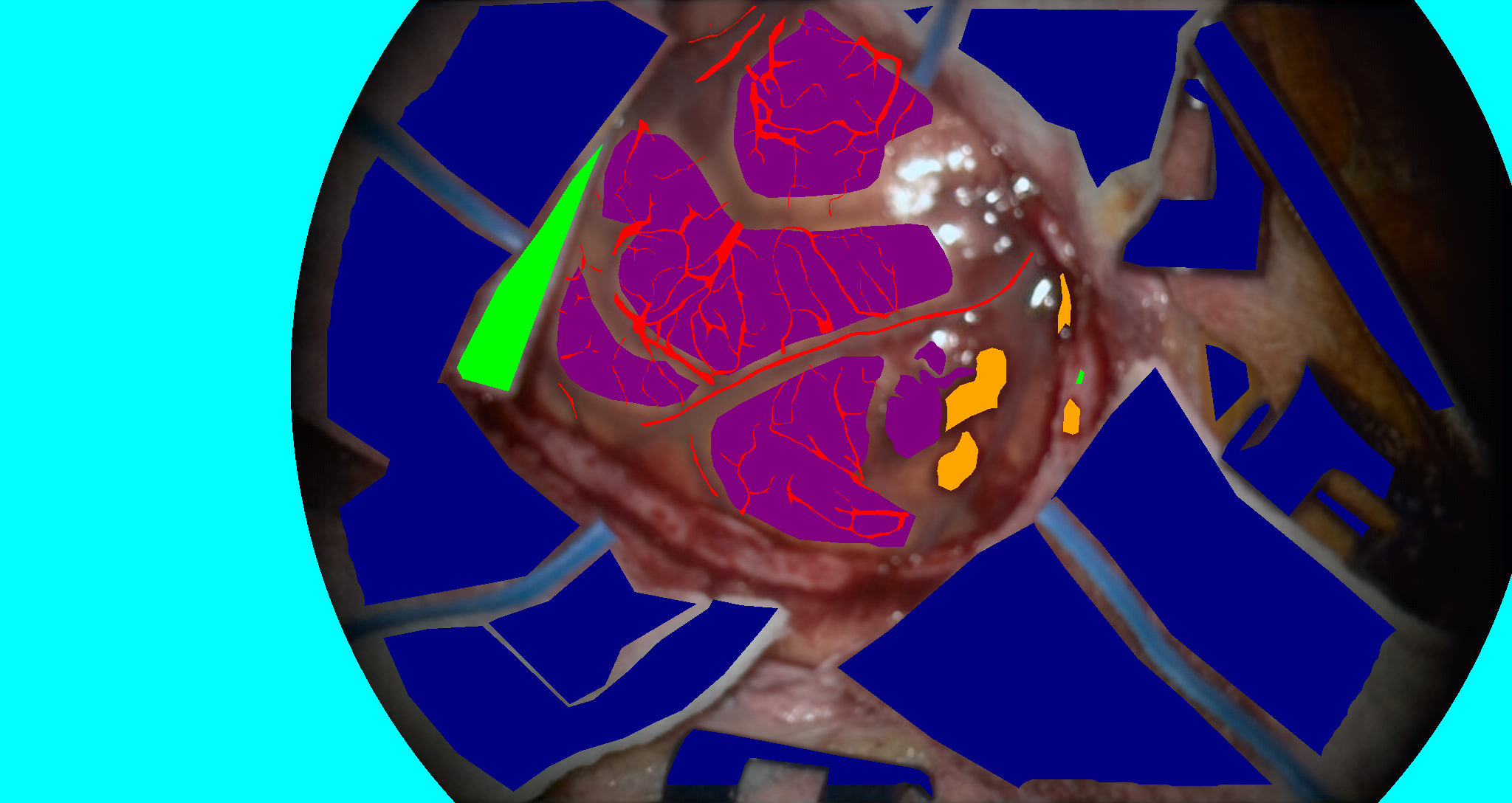}
    & \includegraphics[width=.14\linewidth,valign=m]{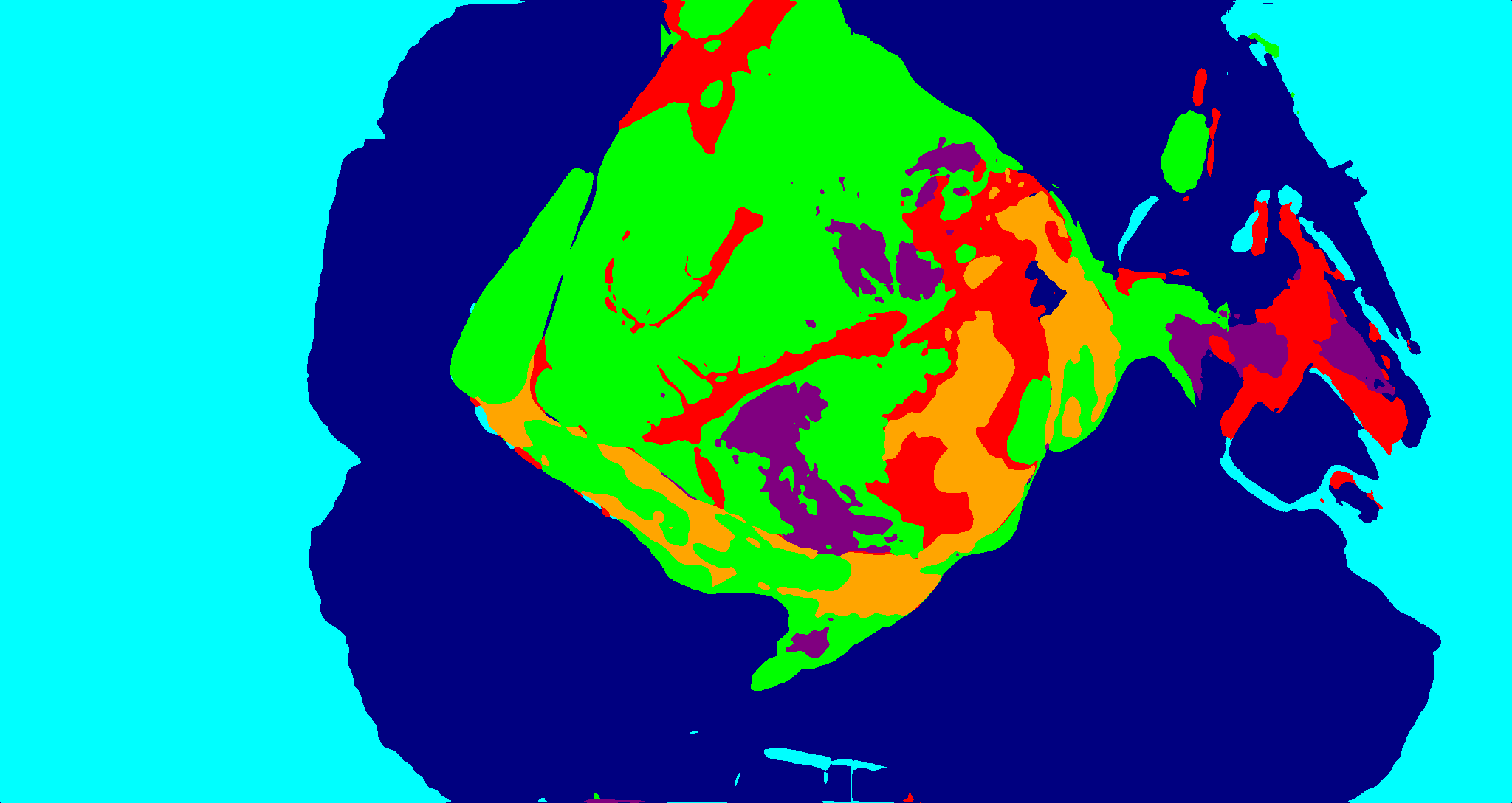}
    & \includegraphics[width=.14\linewidth,valign=m]{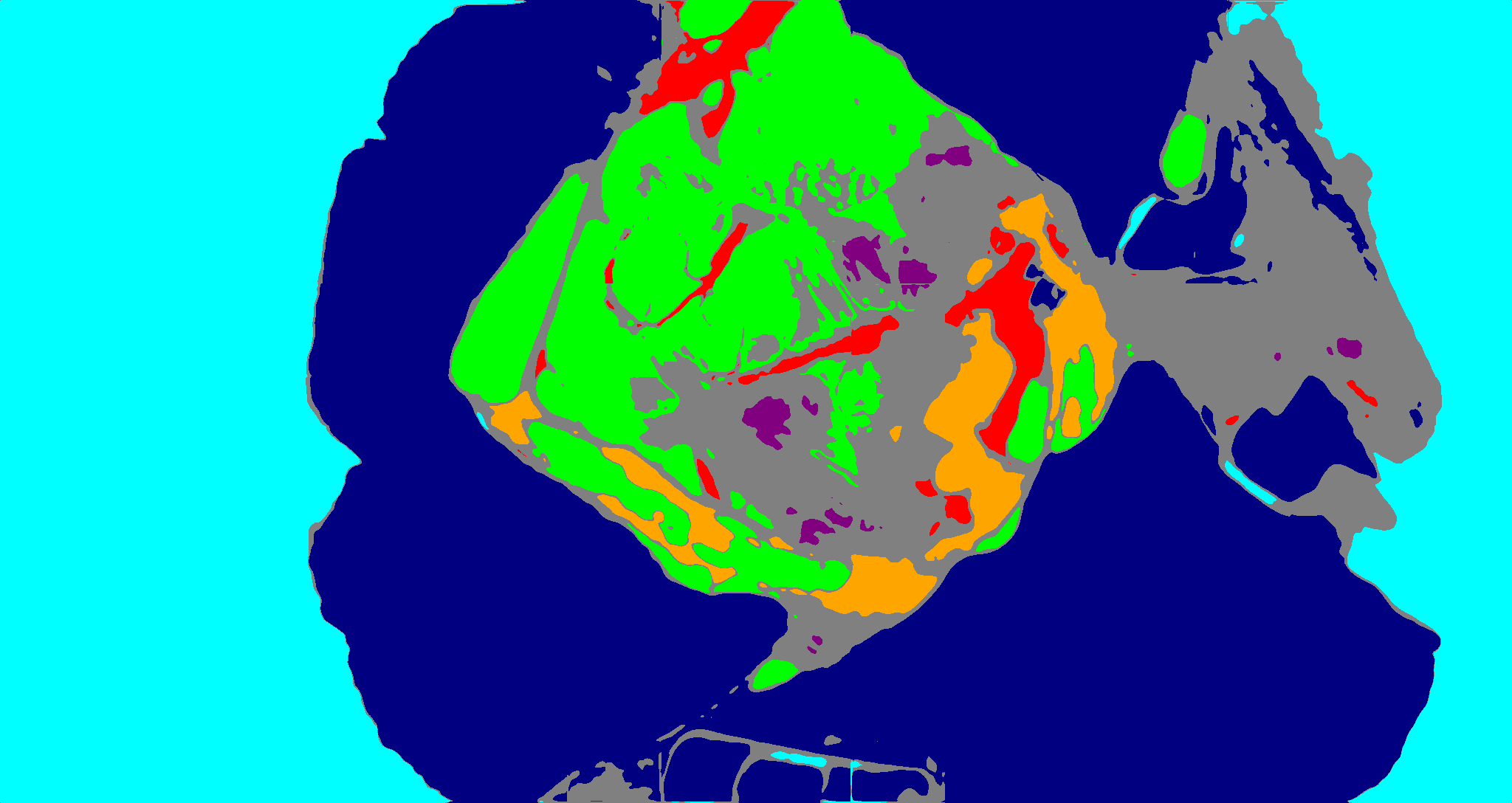}
    & \includegraphics[width=.14\linewidth,valign=m]{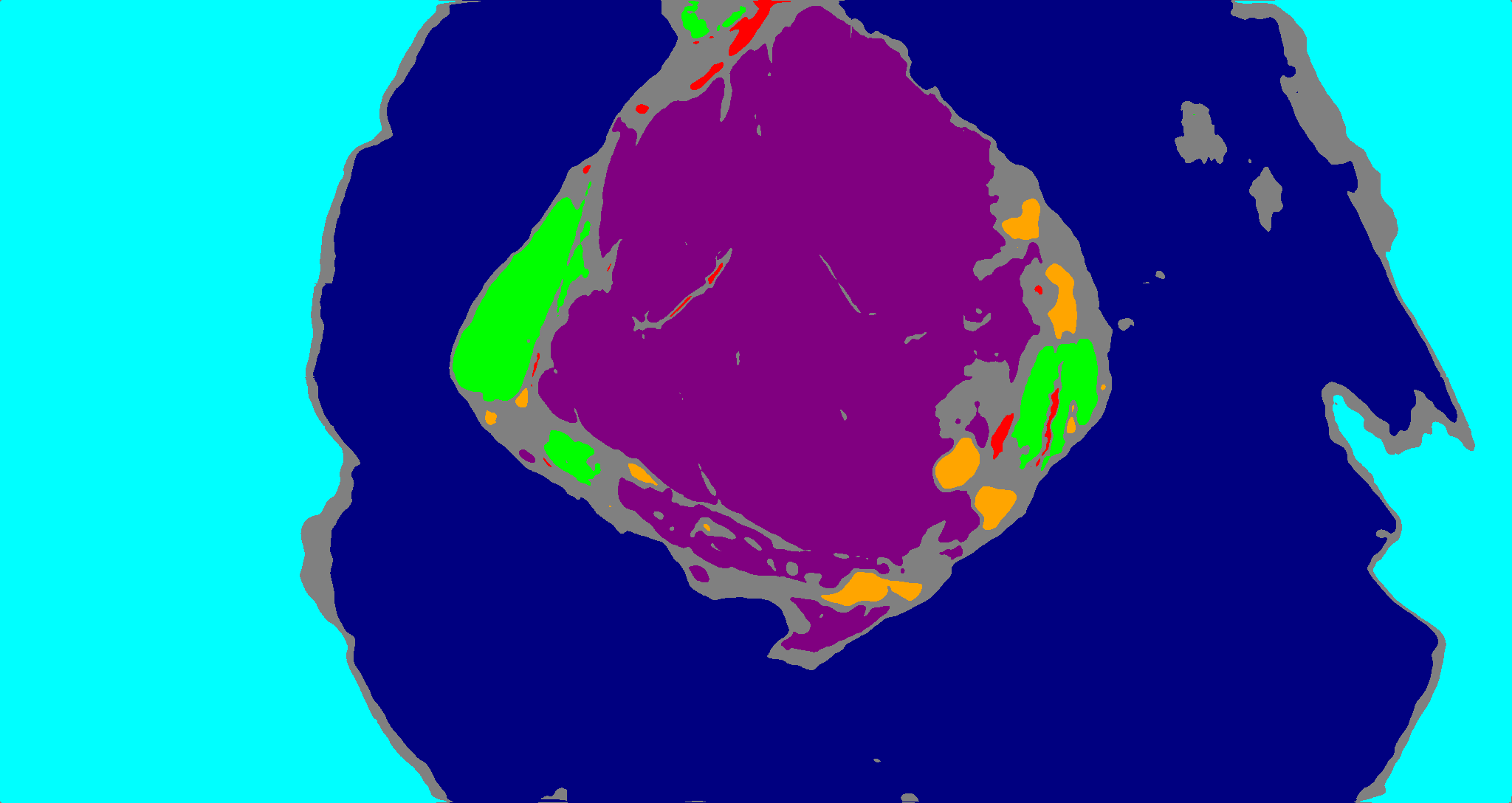}
    & \includegraphics[width=.14\linewidth,valign=m]{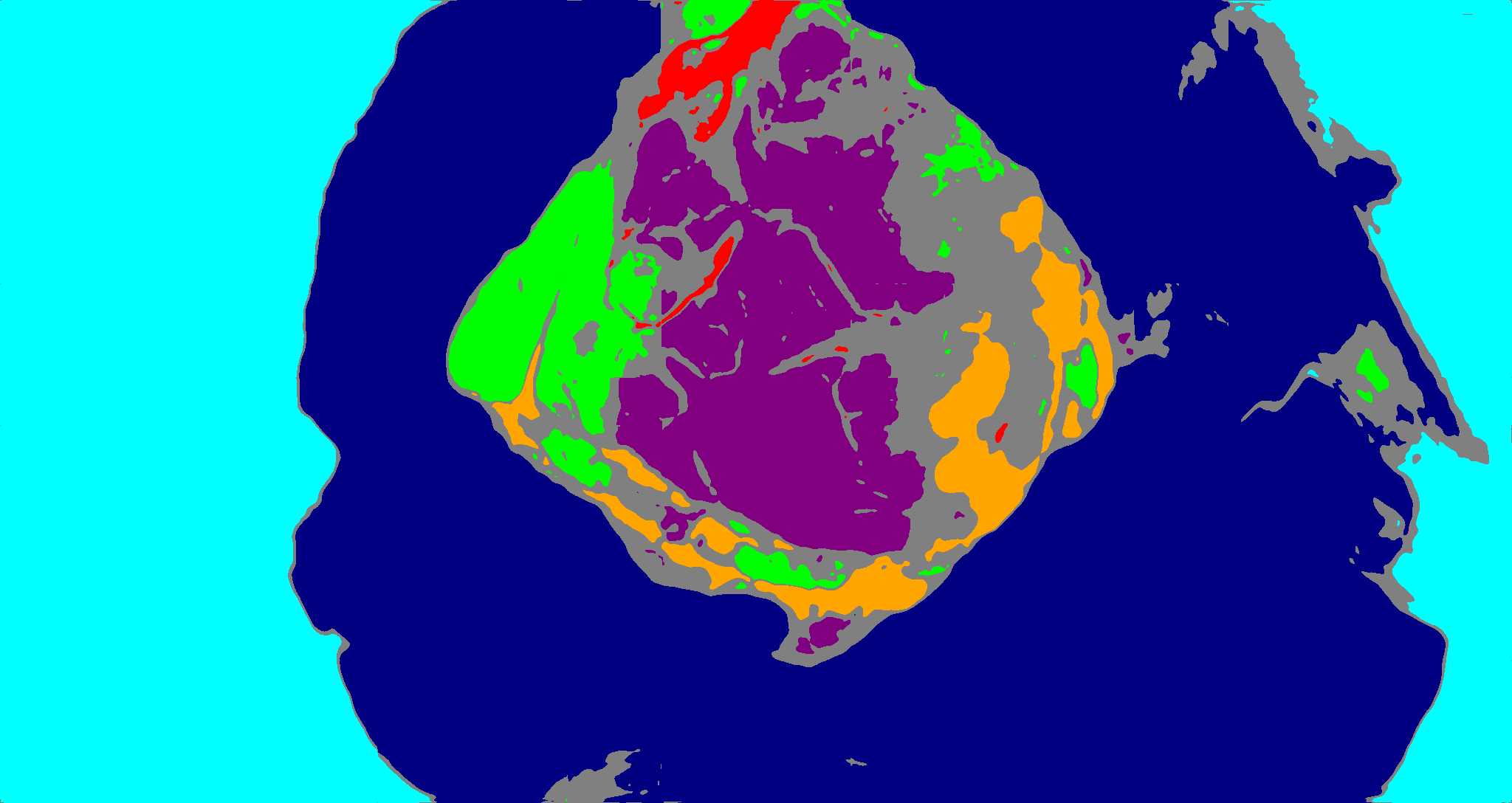}
    & \includegraphics[width=.14\linewidth,valign=m]{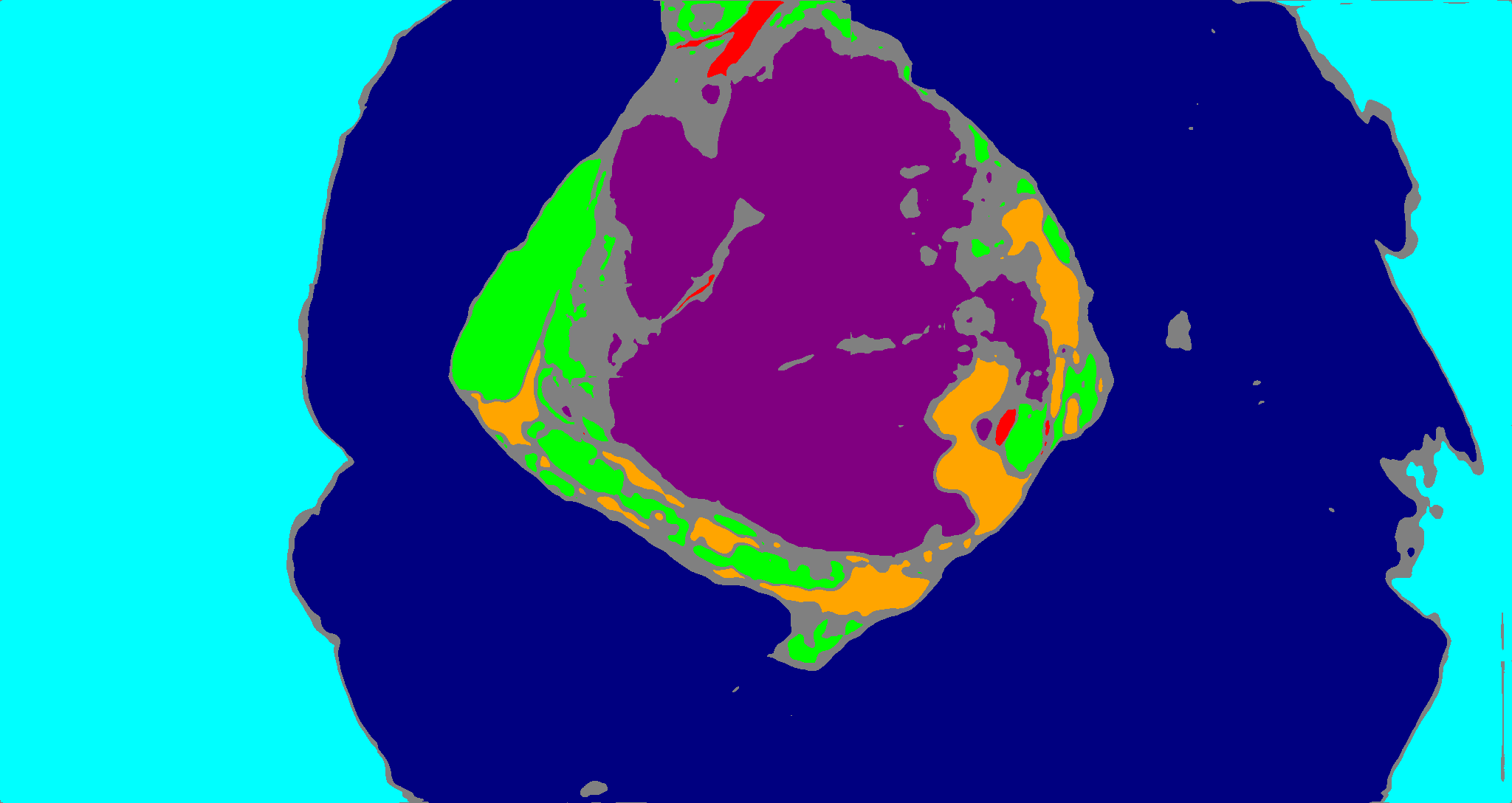}
    \\

    & \includegraphics[width=.14\linewidth,valign=m]{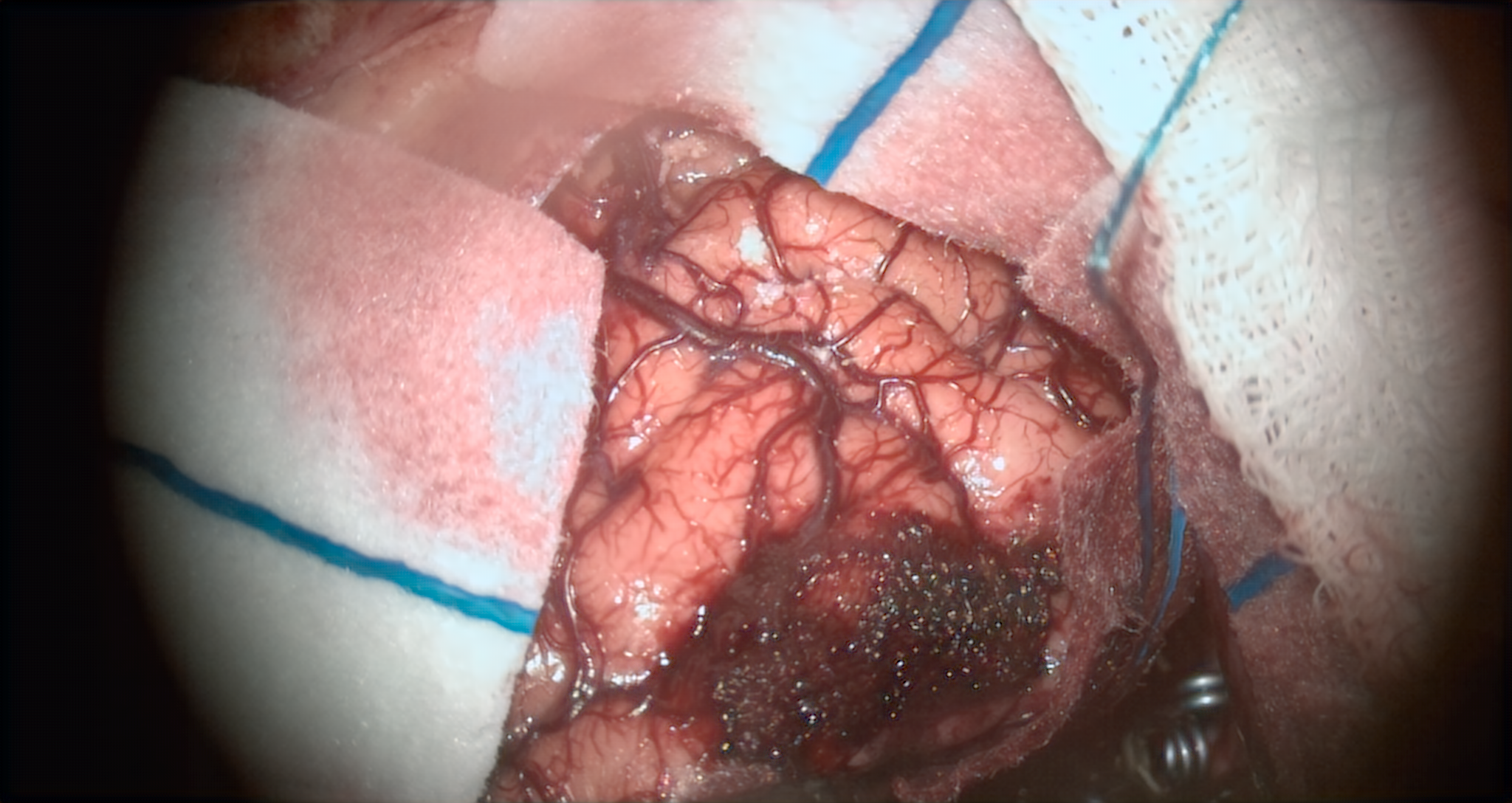}
    & \includegraphics[width=.14\linewidth,valign=m]{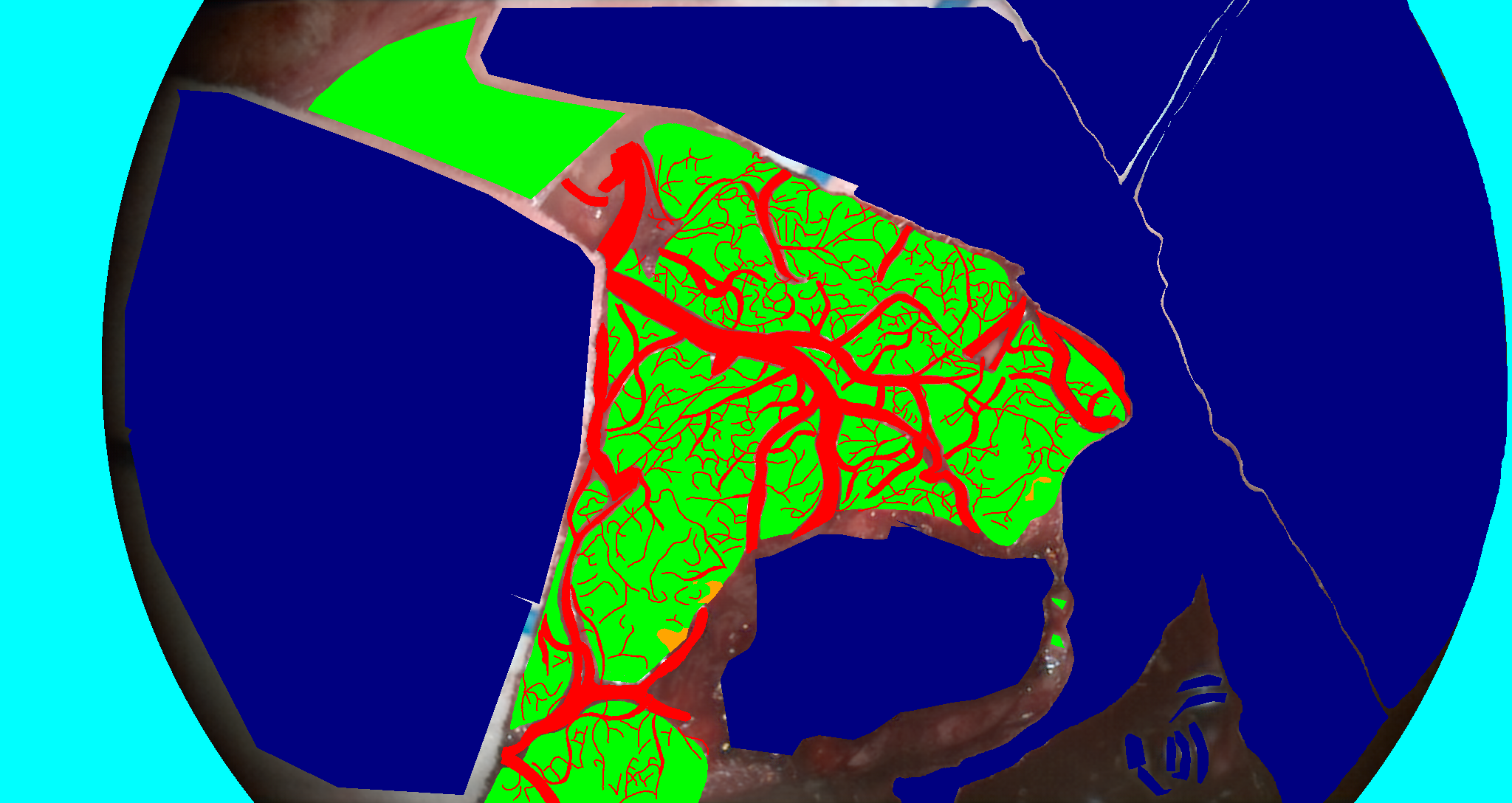}
    & \includegraphics[width=.14\linewidth,valign=m]{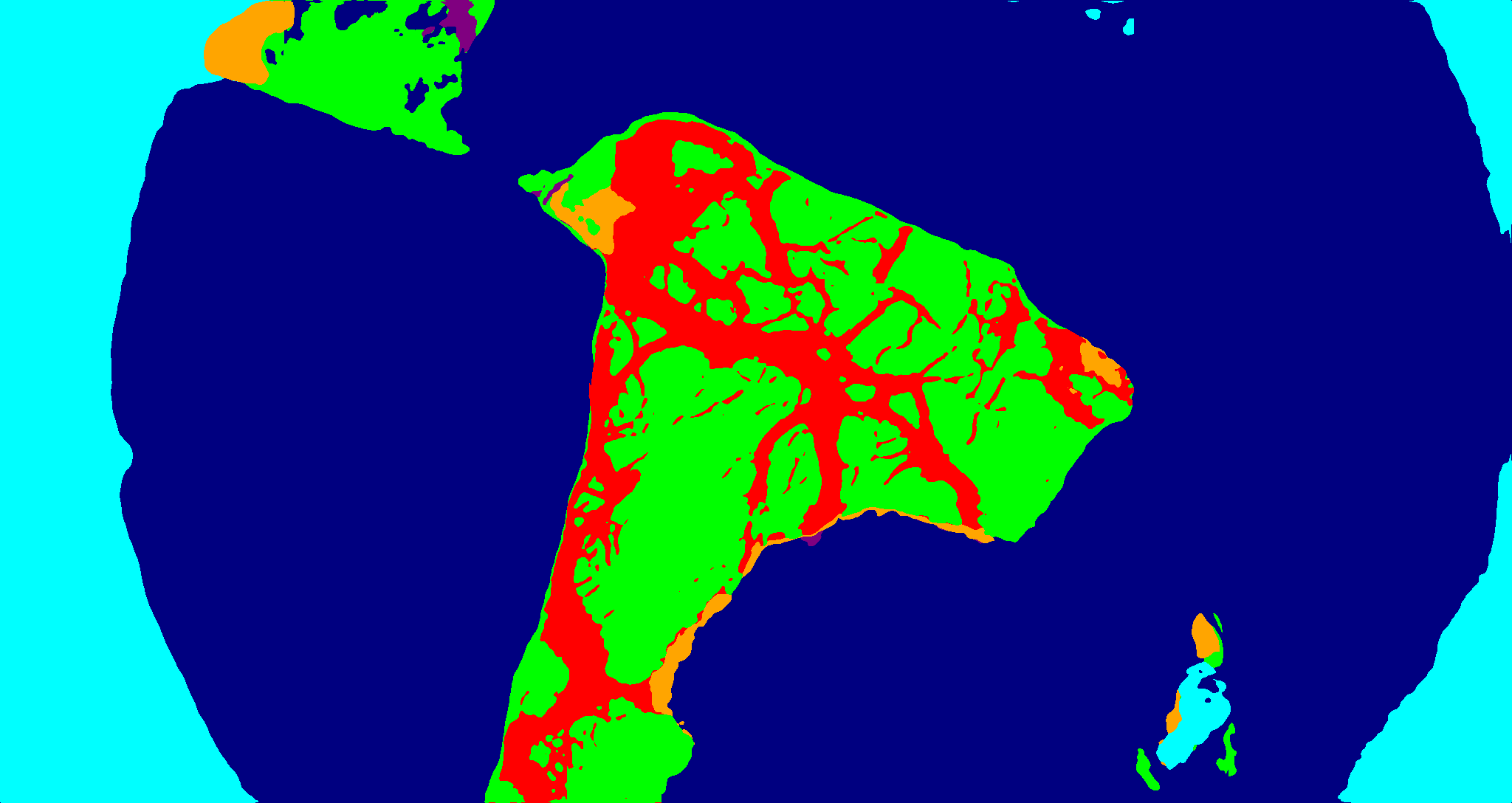}
    & \includegraphics[width=.14\linewidth,valign=m]{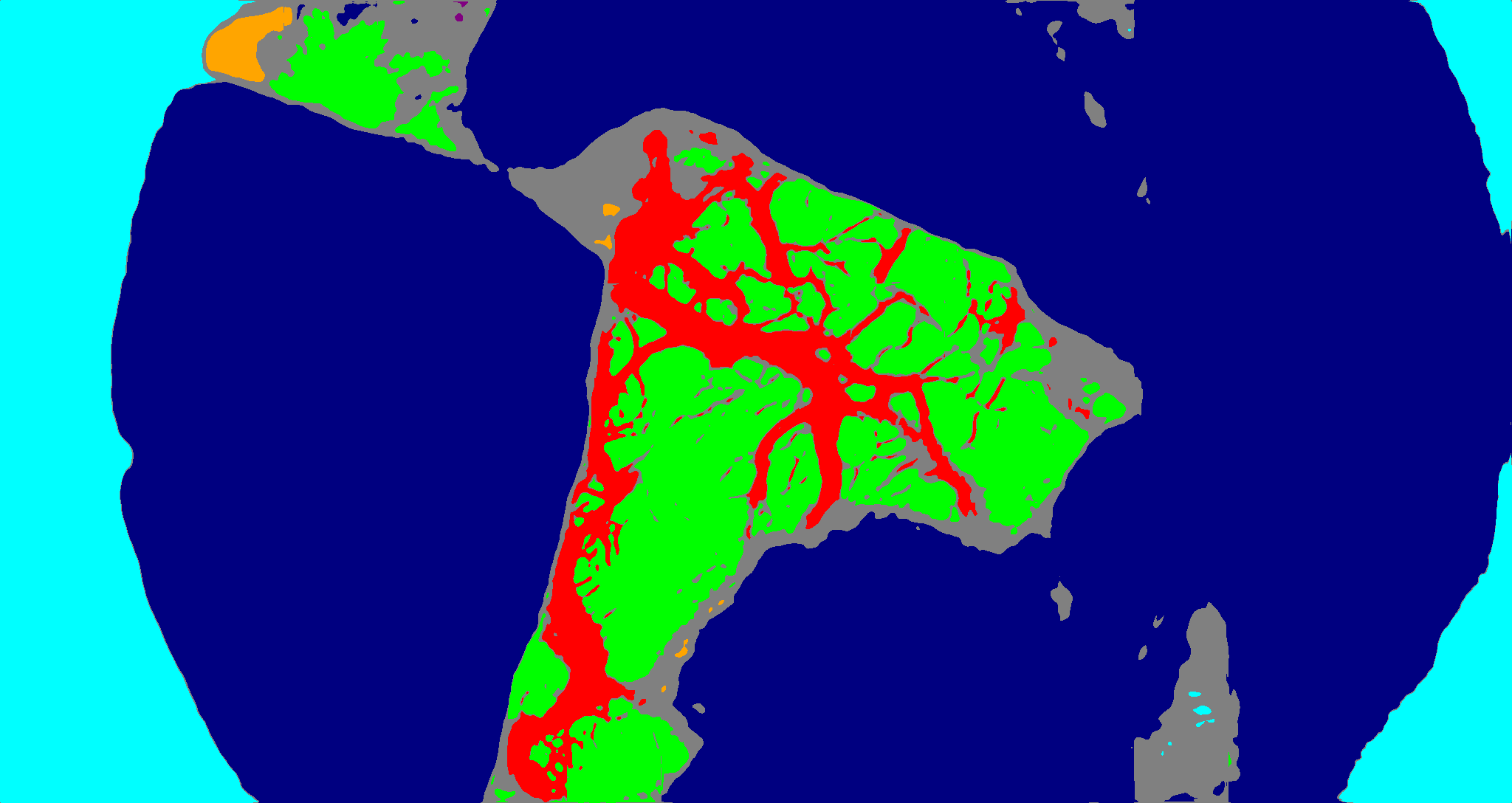}
    & \includegraphics[width=.14\linewidth,valign=m]{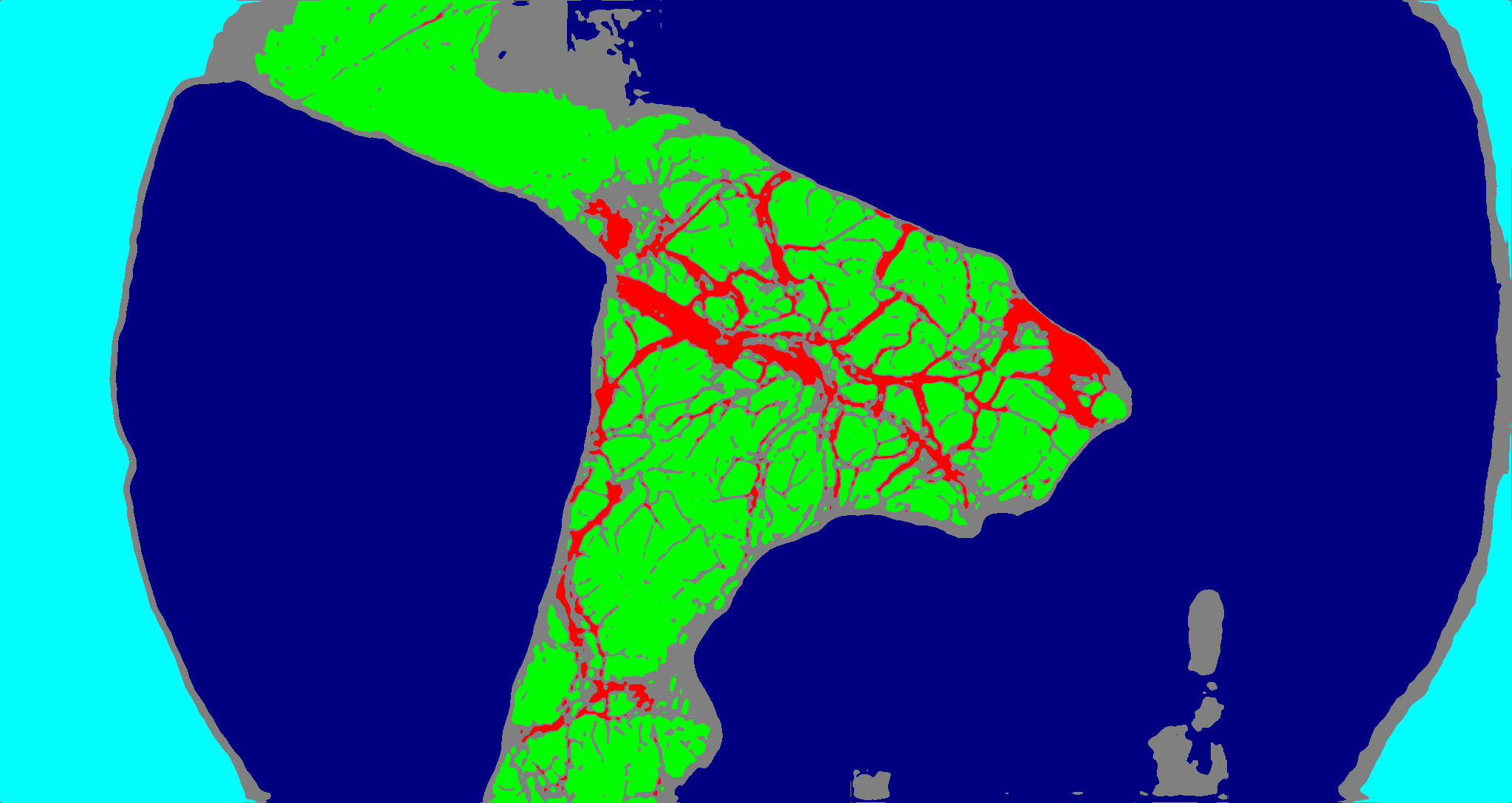}
    & \includegraphics[width=.14\linewidth,valign=m]{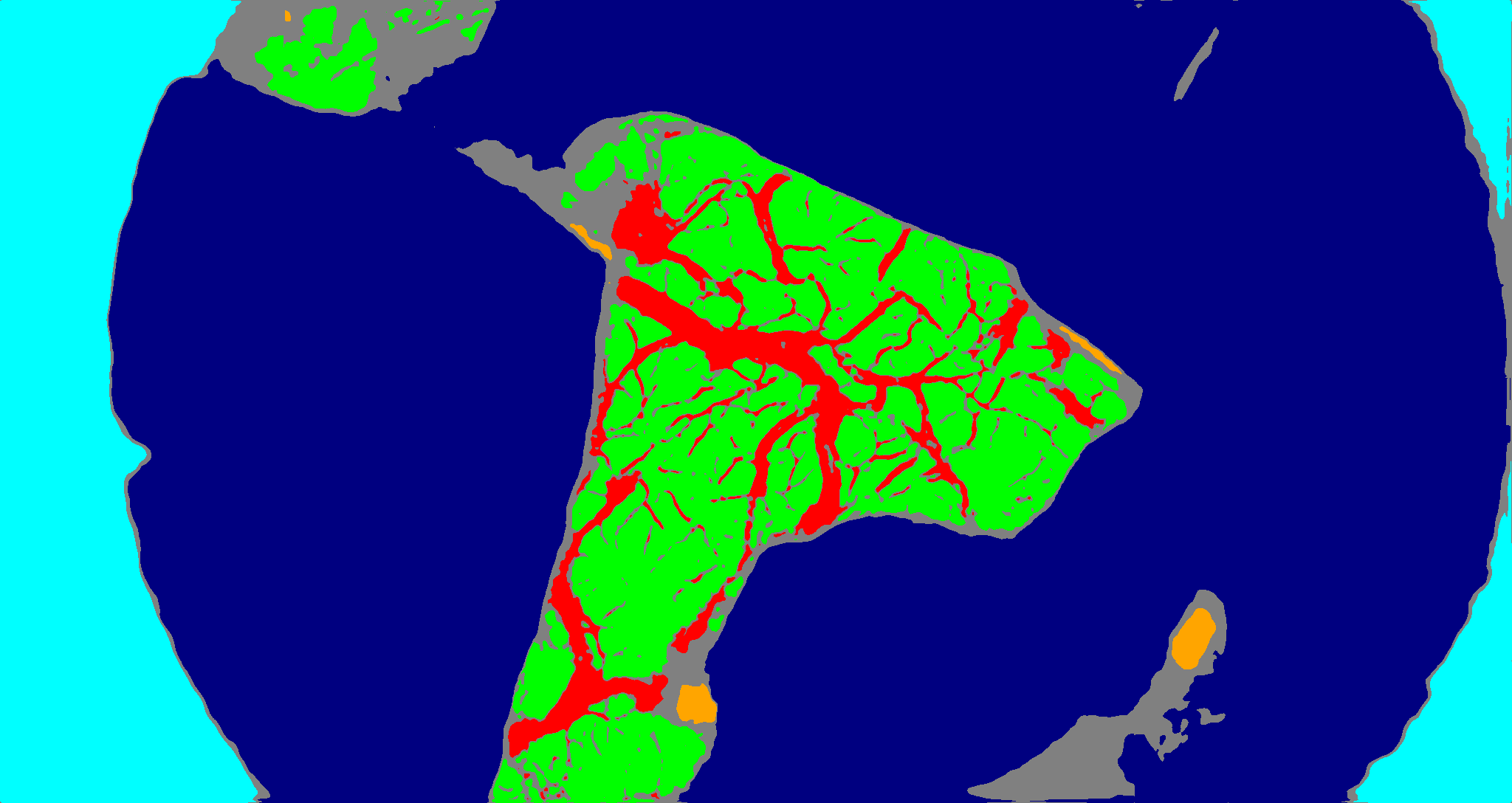}
    & \includegraphics[width=.14\linewidth,valign=m]{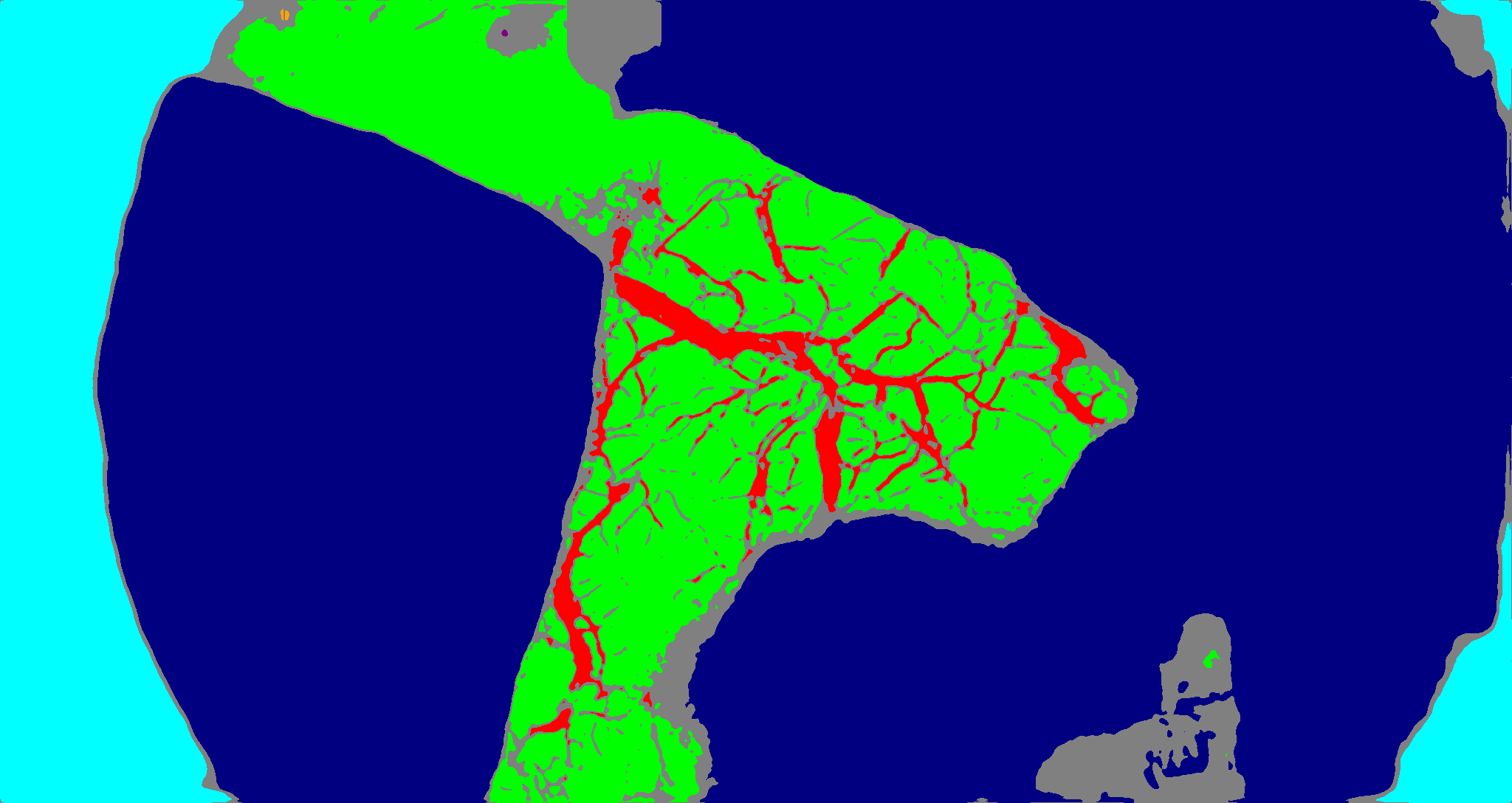}
    \\

    & \includegraphics[width=.14\linewidth,valign=m]{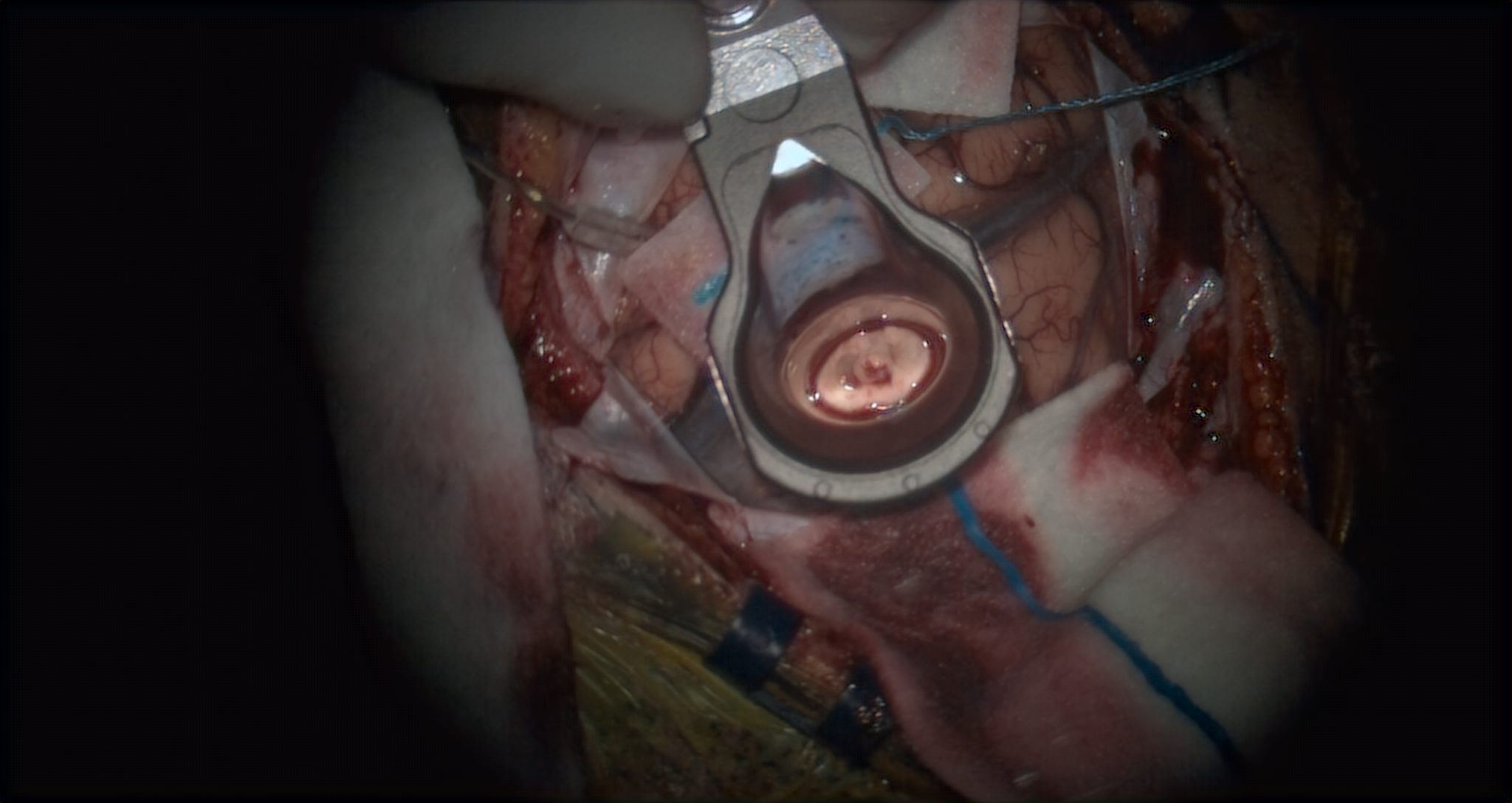}
    & \includegraphics[width=.14\linewidth,valign=m]{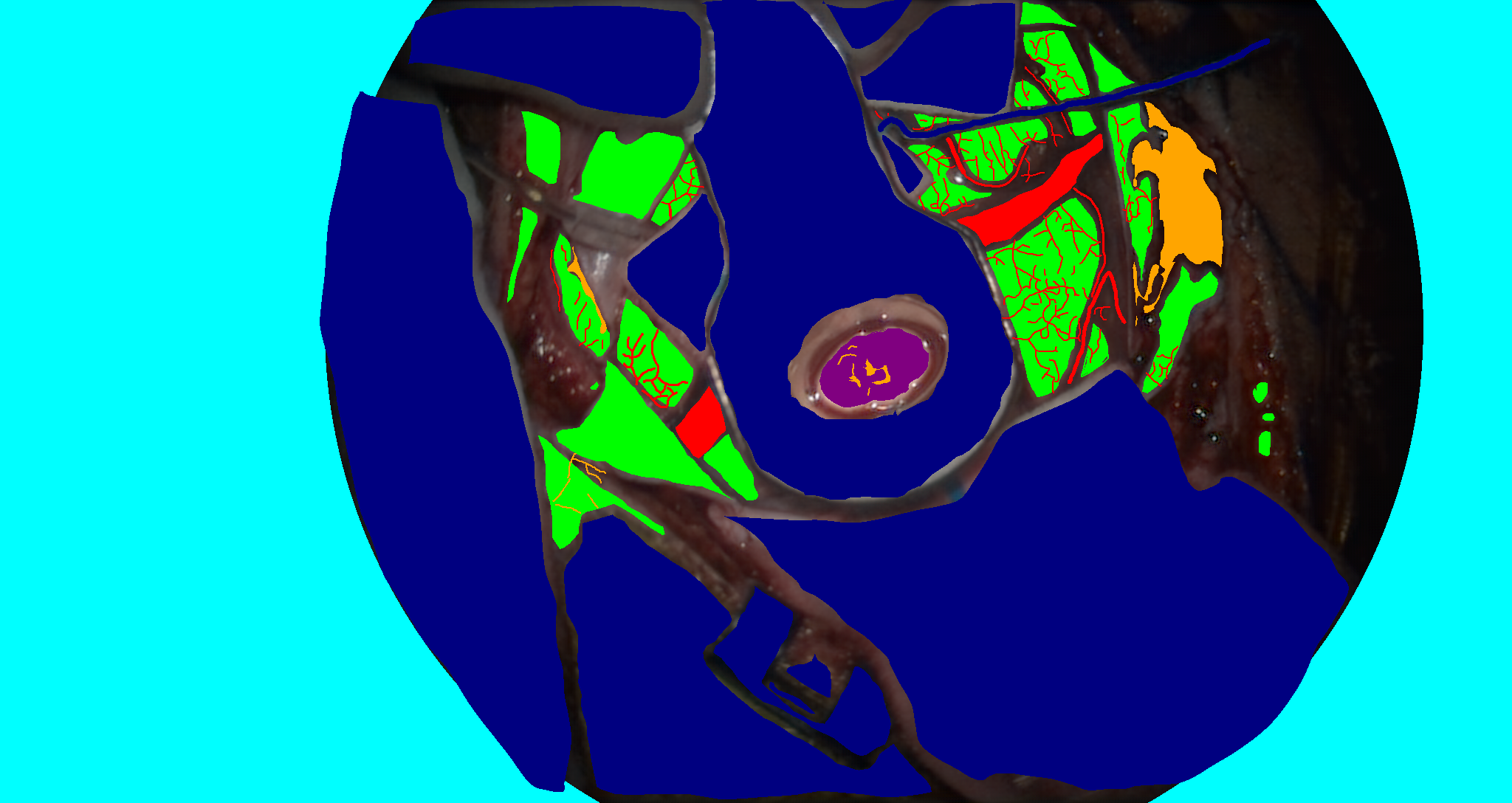}
    & \includegraphics[width=.14\linewidth,valign=m]{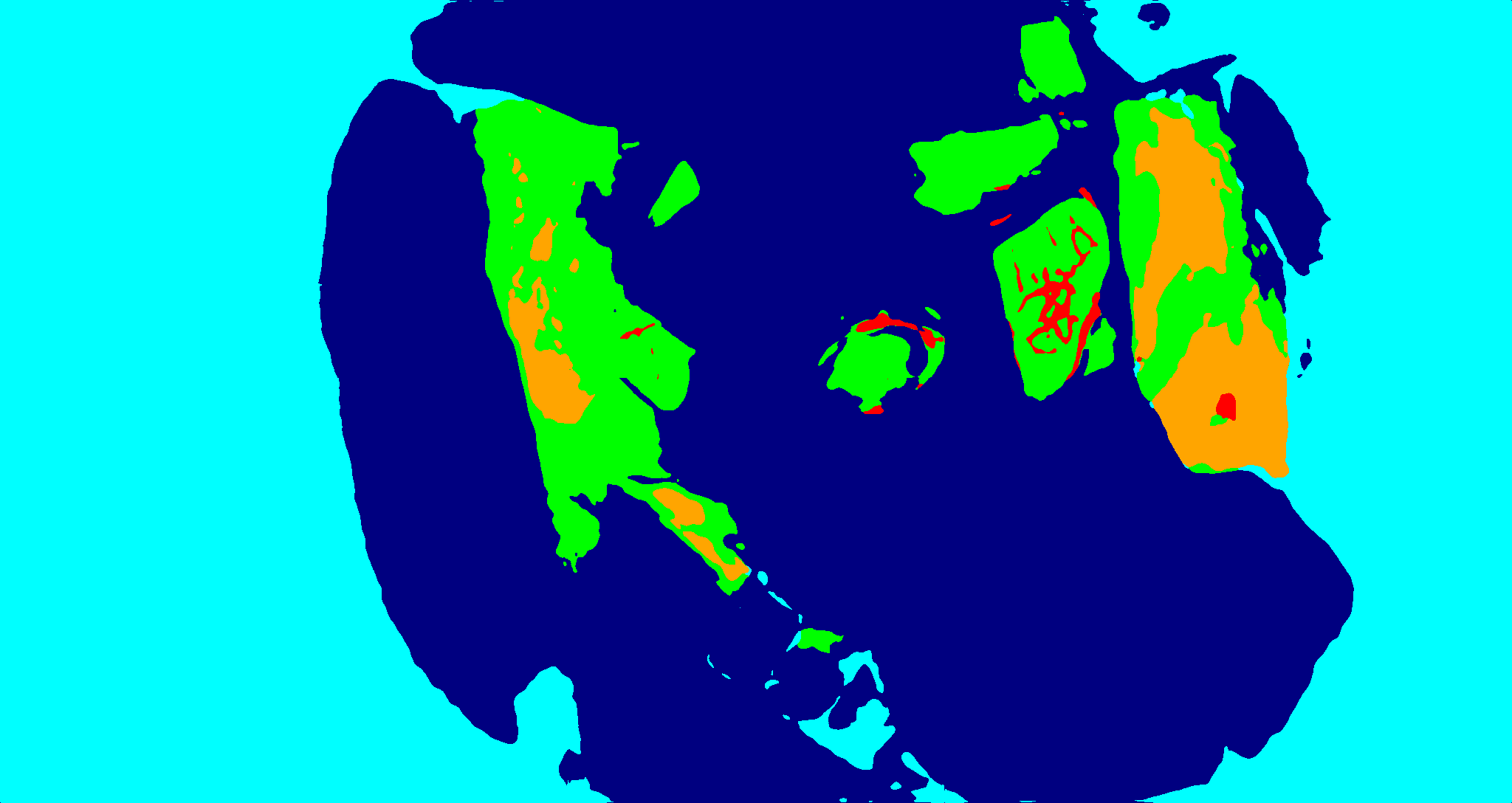}
    & \includegraphics[width=.14\linewidth,valign=m]{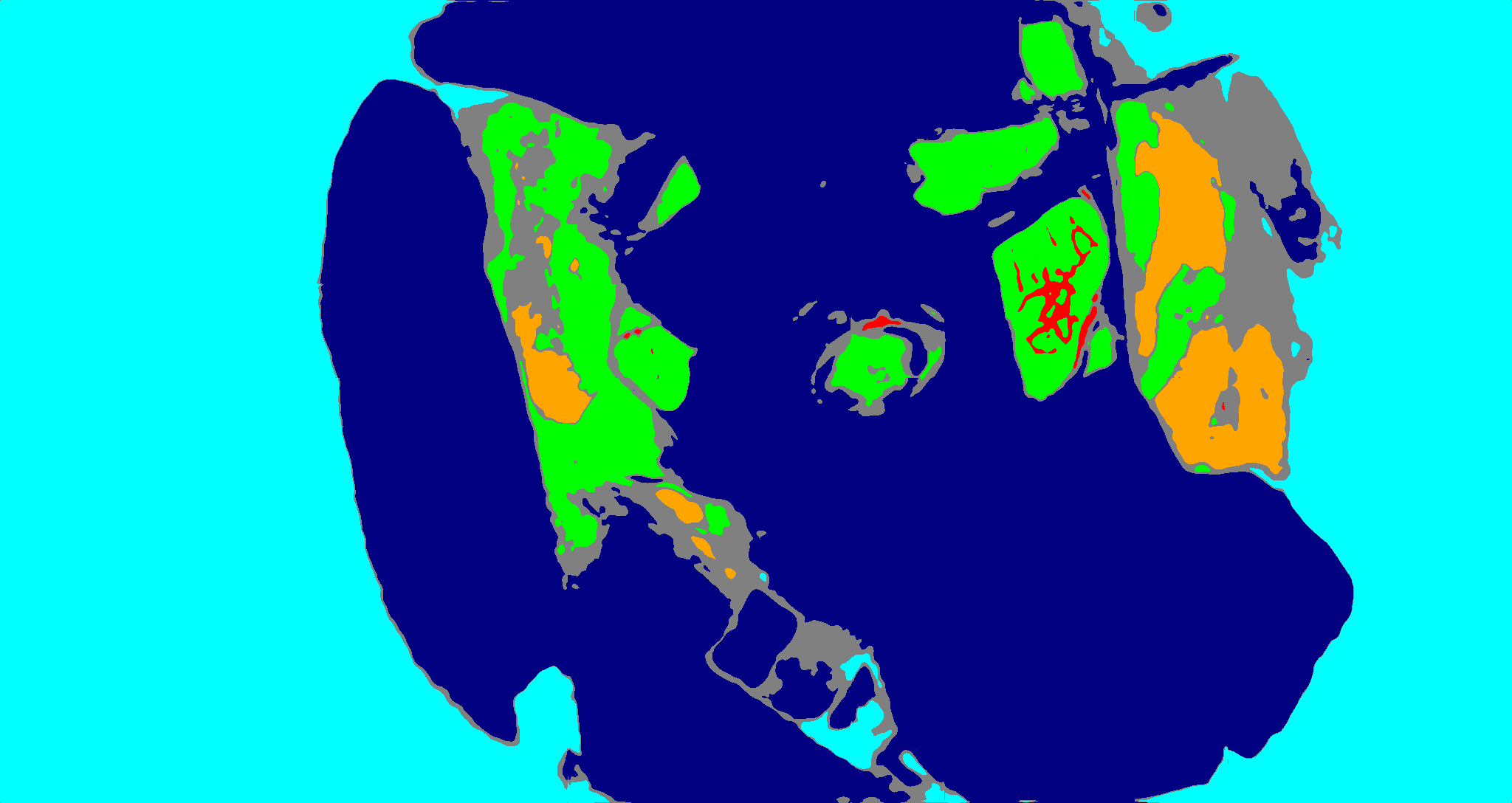}
    & \includegraphics[width=.14\linewidth,valign=m]{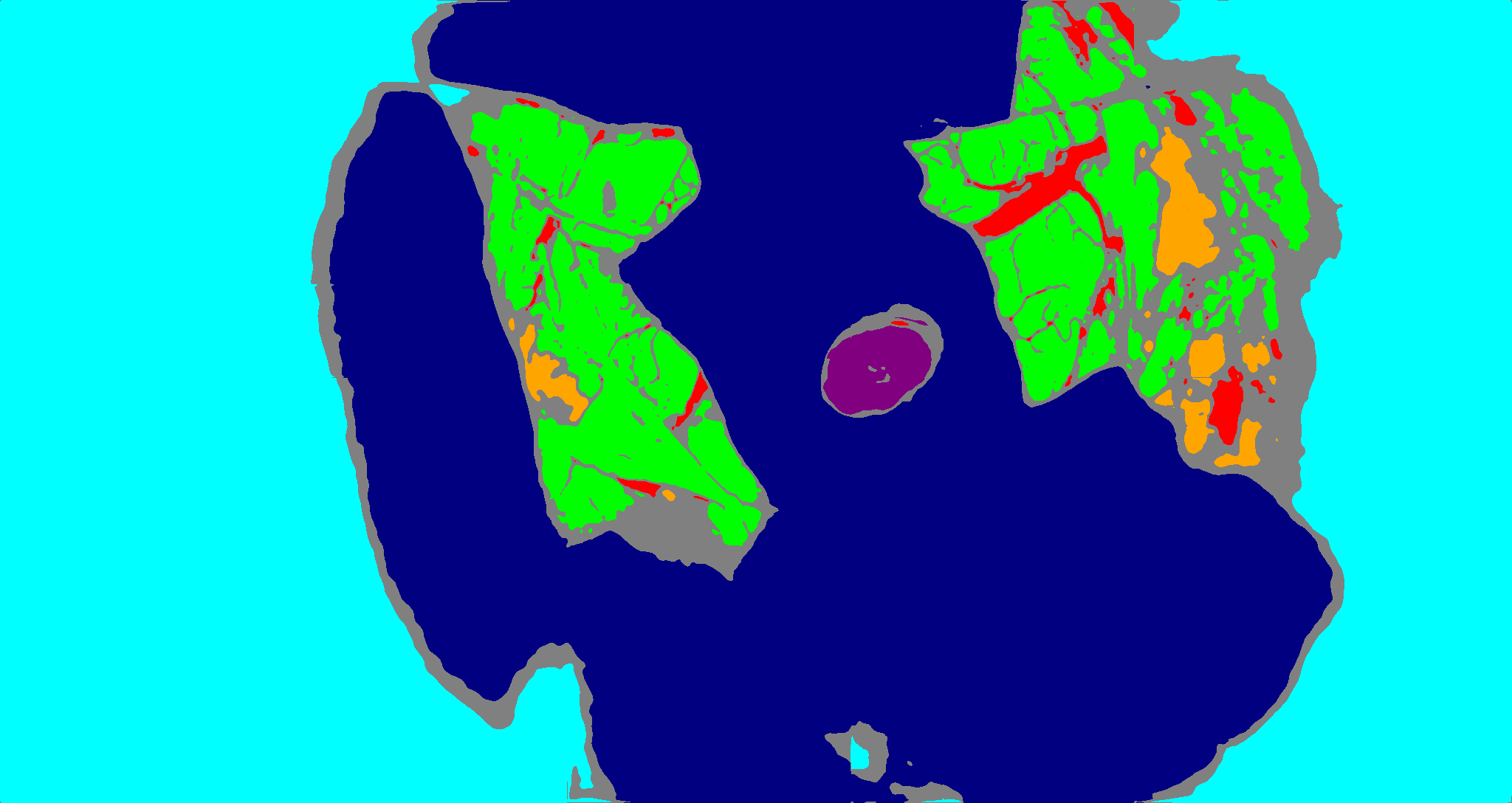}
    & \includegraphics[width=.14\linewidth,valign=m]{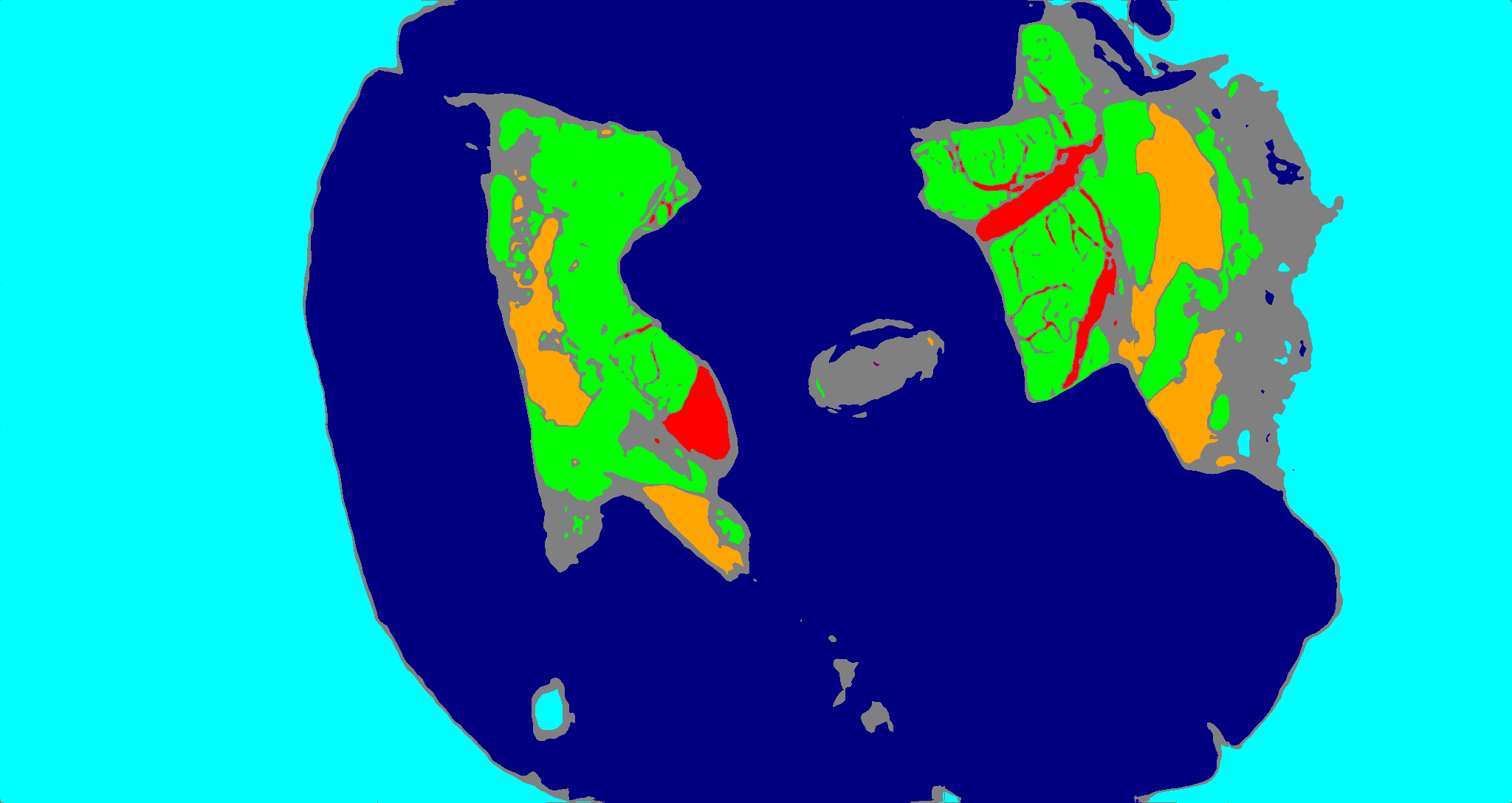}
    & \includegraphics[width=.14\linewidth,valign=m]{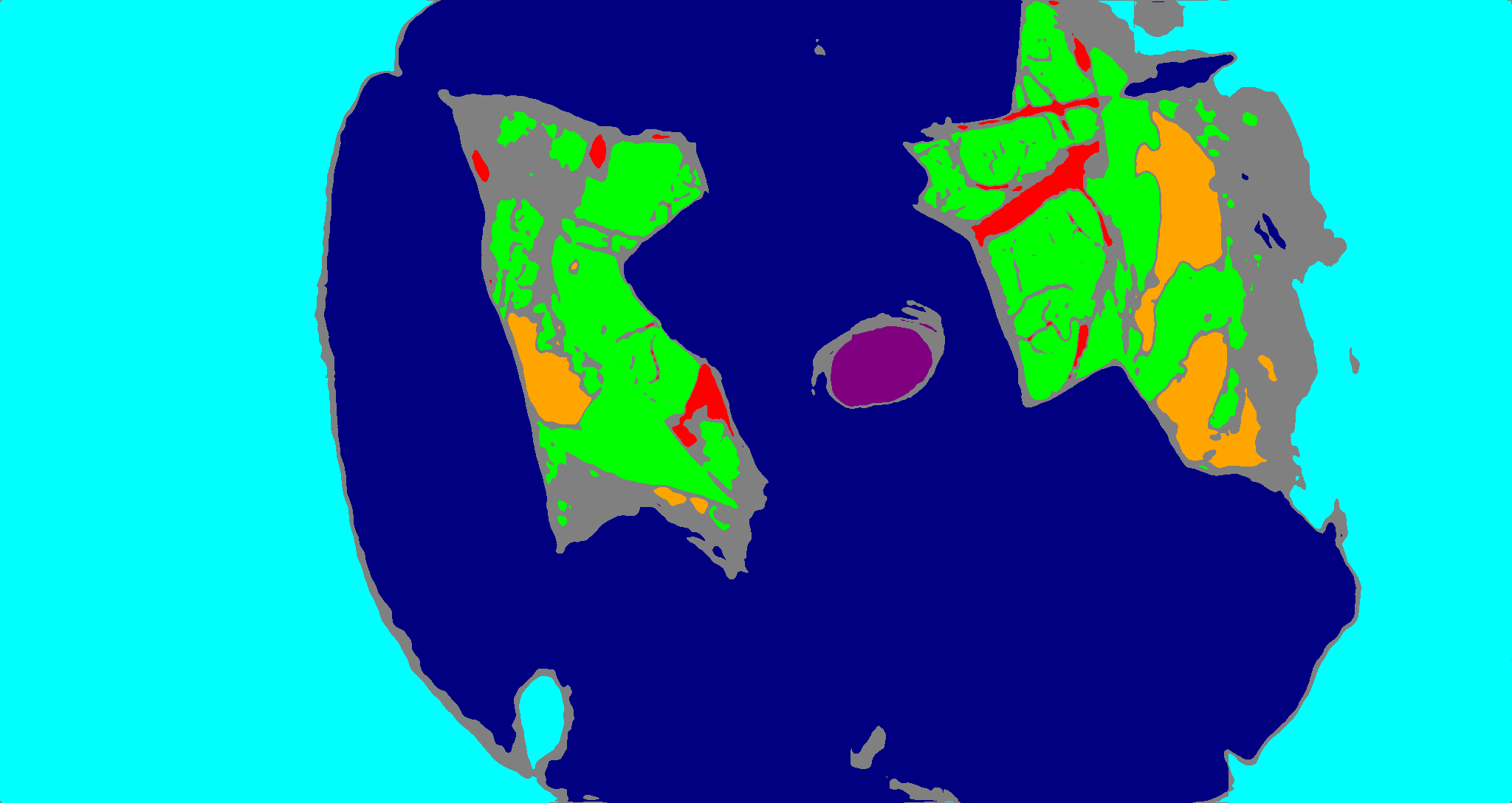}
    \\

    & \includegraphics[width=.14\linewidth,valign=m]{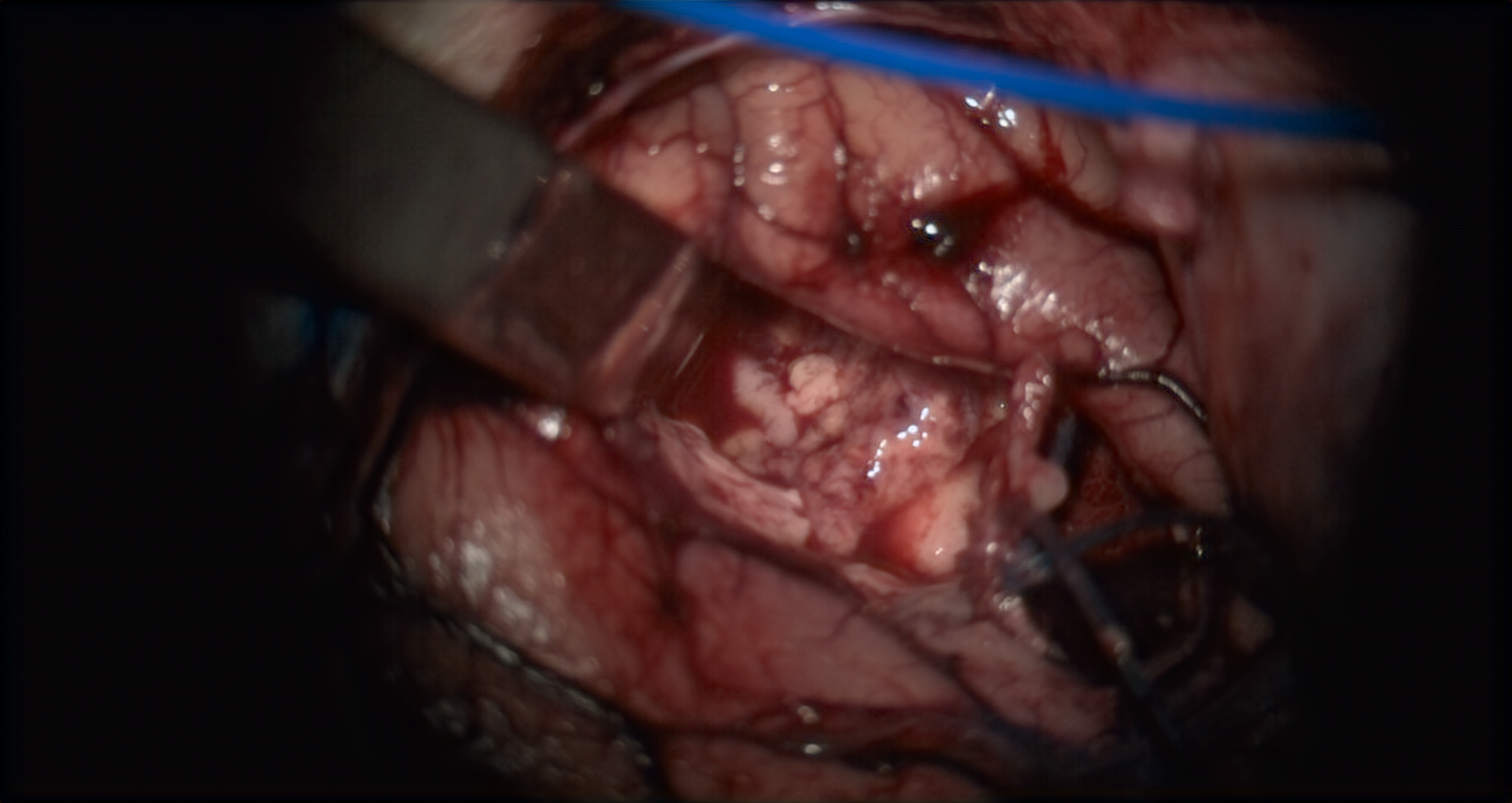}
    & \includegraphics[width=.14\linewidth,valign=m]{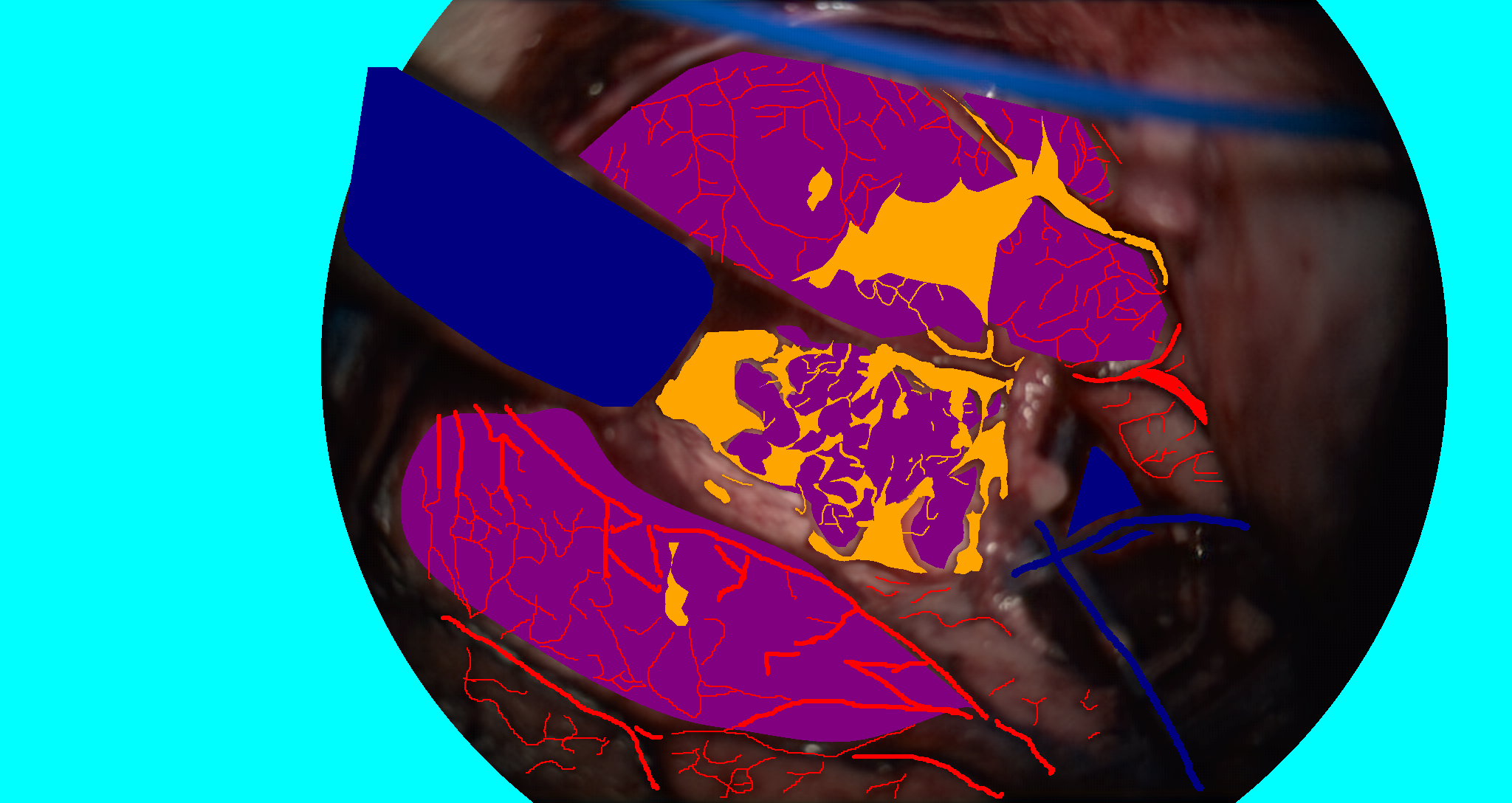}
    & \includegraphics[width=.14\linewidth,valign=m]{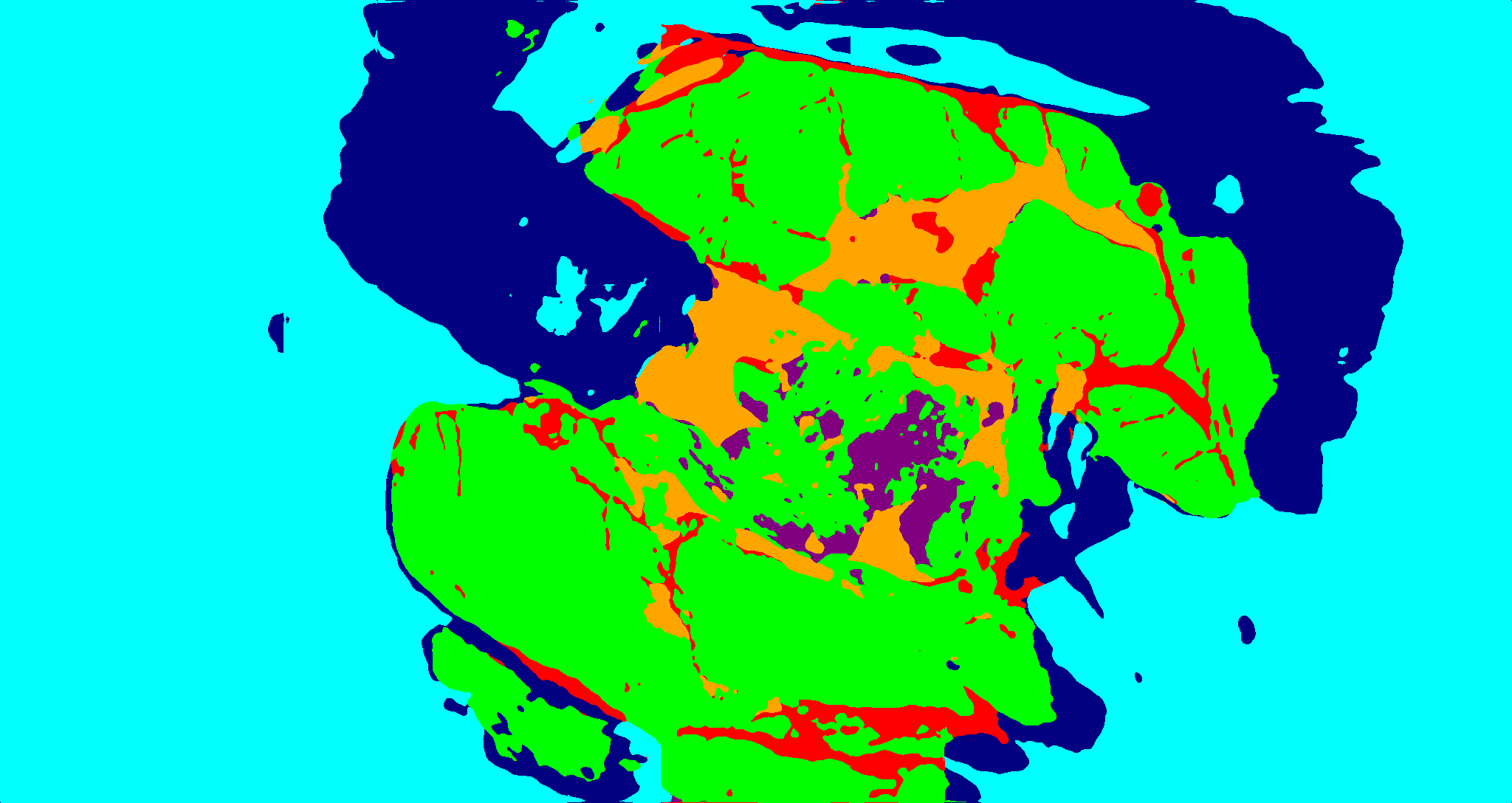}
    & \includegraphics[width=.14\linewidth,valign=m]{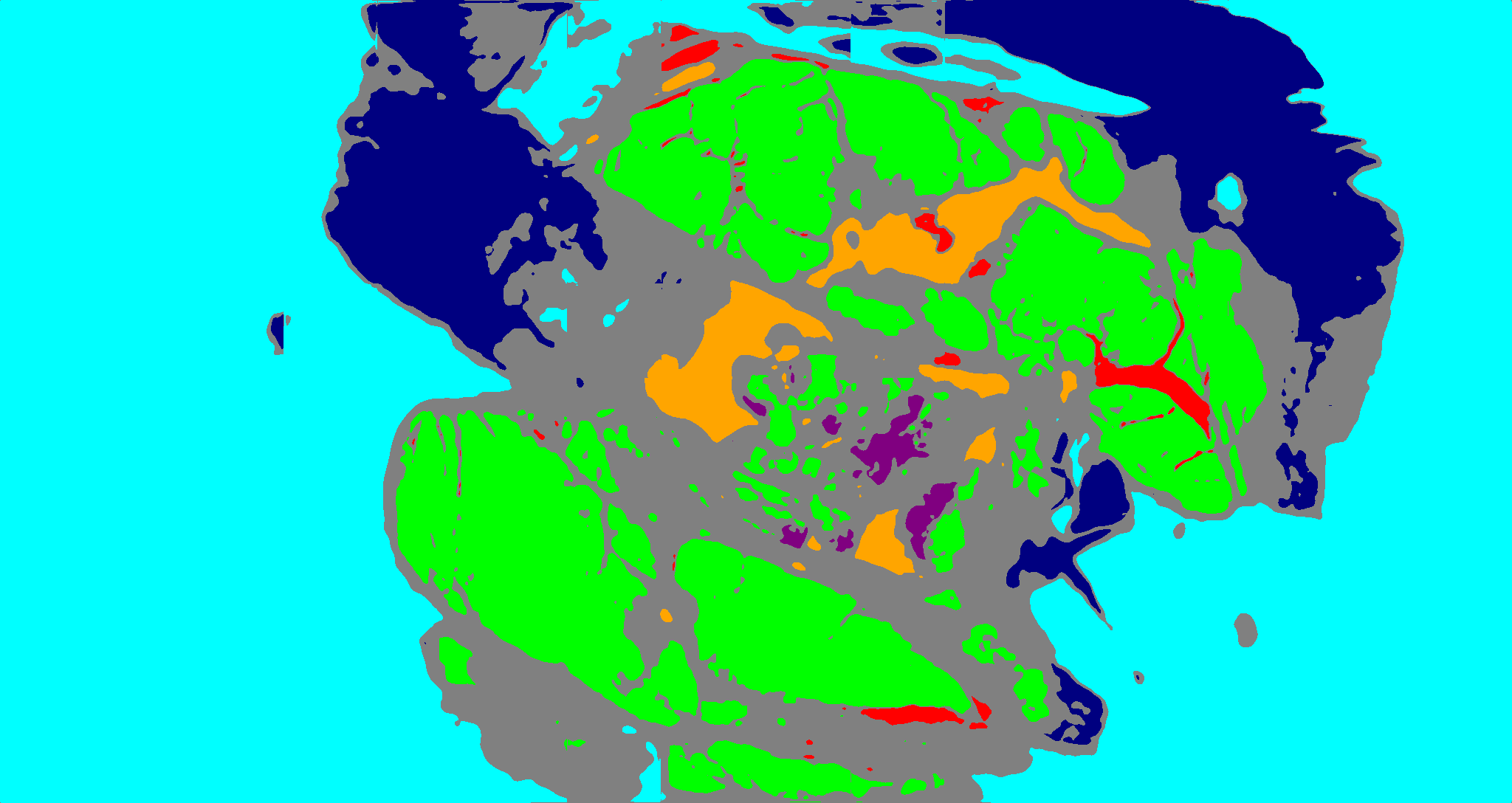}
    & \includegraphics[width=.14\linewidth,valign=m]{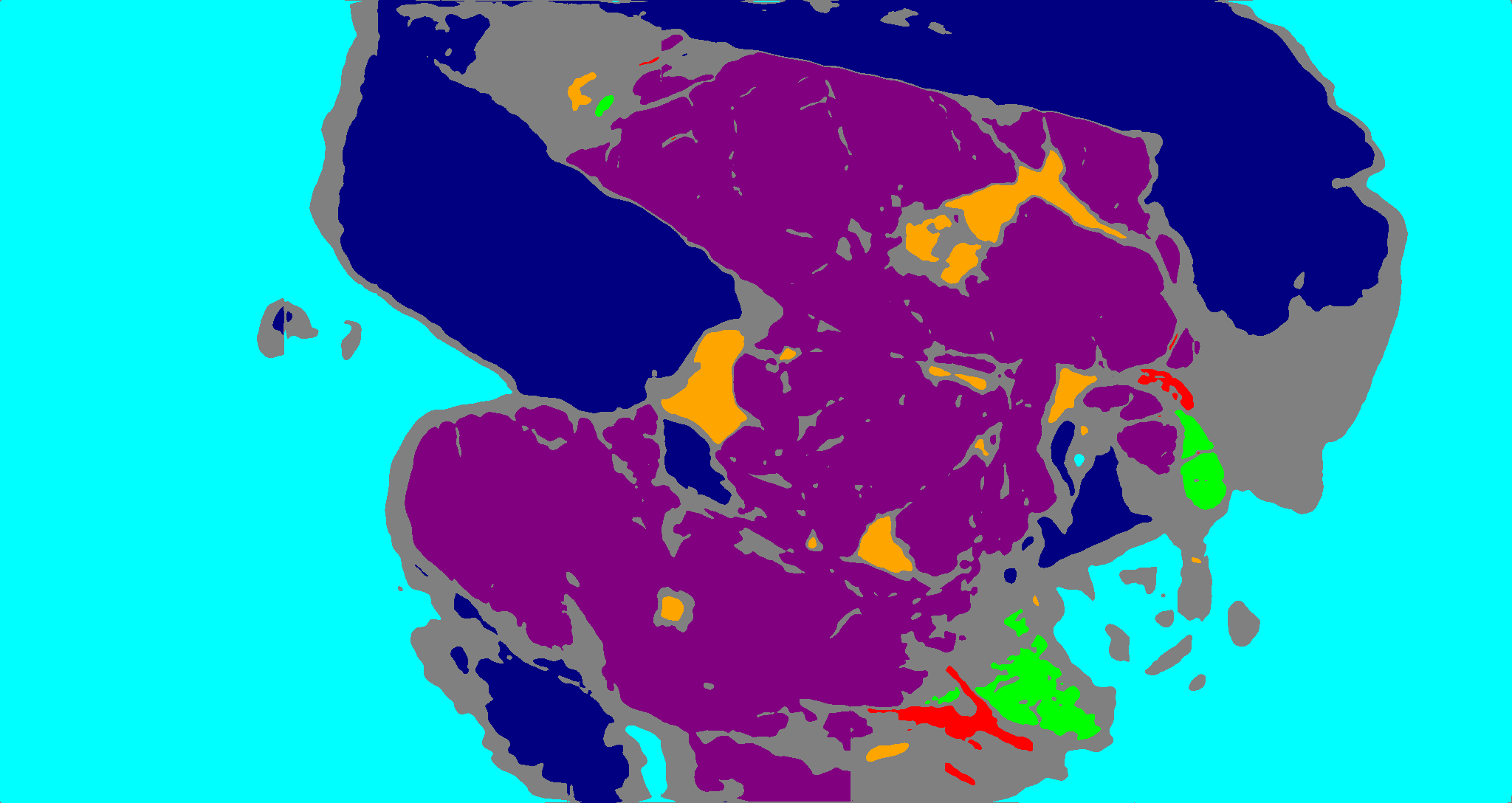}
    & \includegraphics[width=.14\linewidth,valign=m]{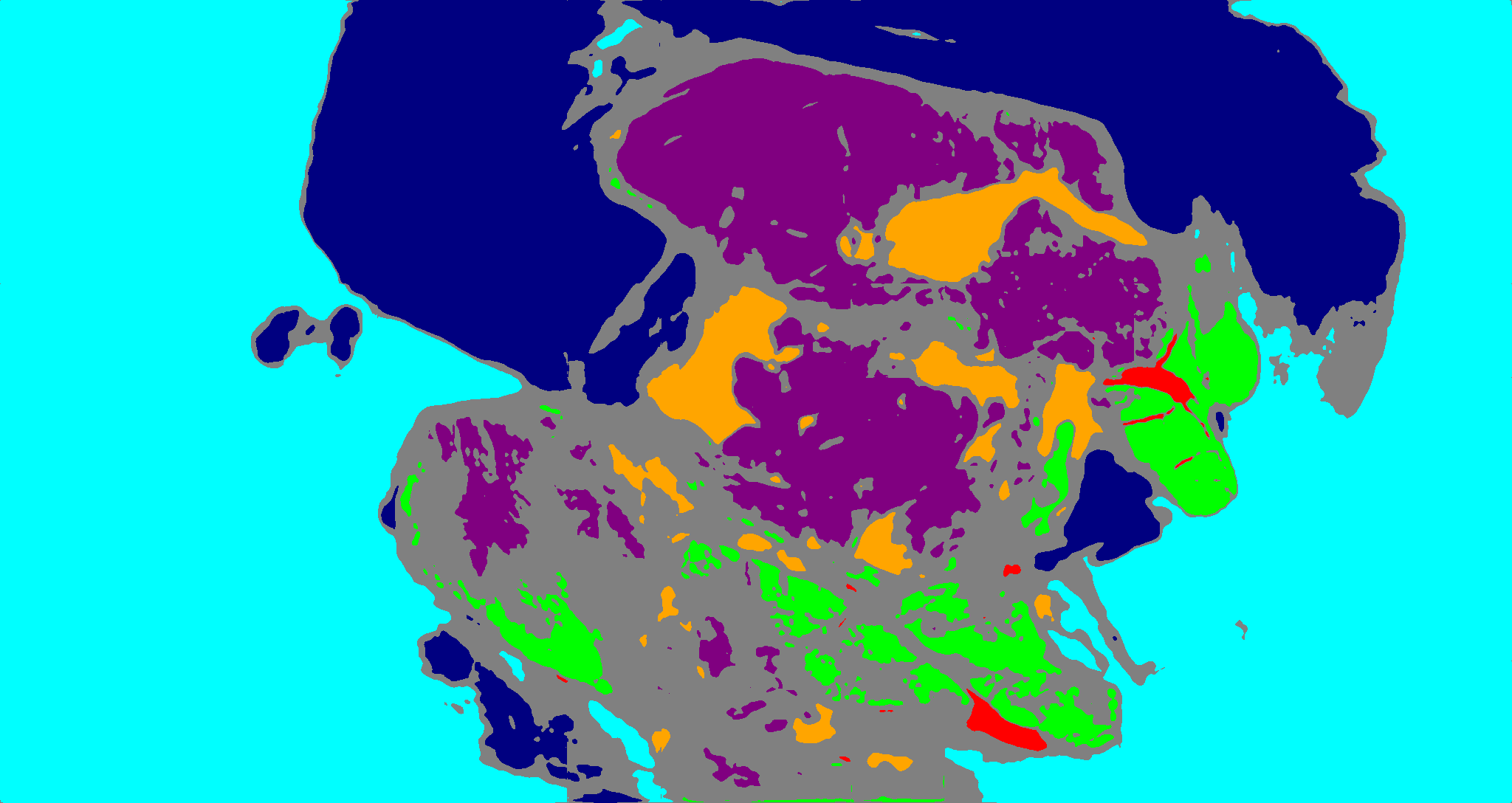}
    & \includegraphics[width=.14\linewidth,valign=m]{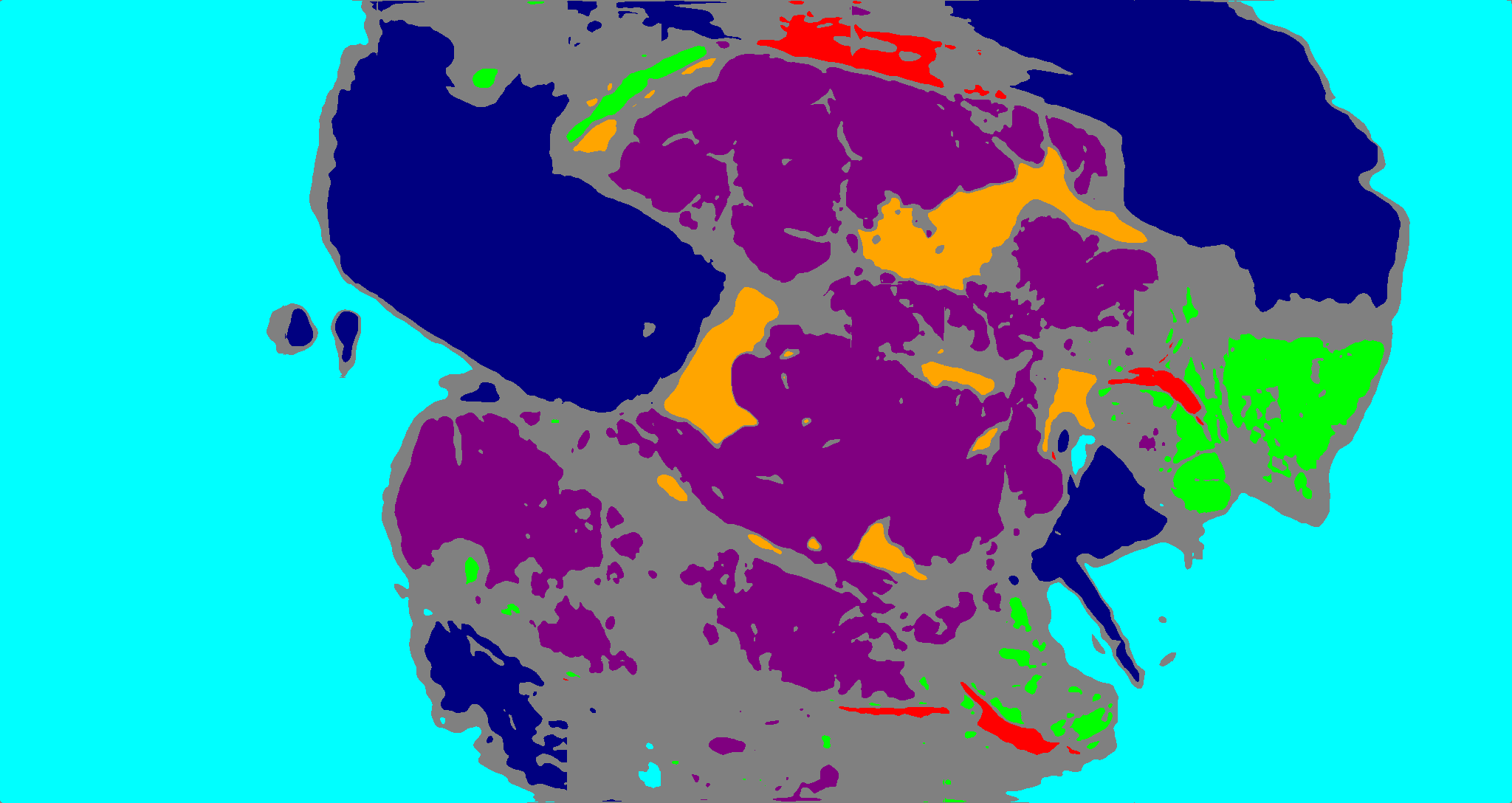}
    \\
    
    & \multicolumn{7}{c}{\includegraphics[width=0.98\linewidth]{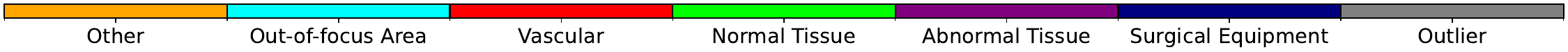}}
\end{tabular}
\caption{Qualitative result on top-level classes. We show the result of same image using different methods at confidence threshold $\tau_m$. Baseline results at $\tau_0=0$ are added to represent result without outlier detection.}
\label{fig:qualitative_results_all_classes}
\end{figure}
\figref{fig:qualitative_results_all_classes} presents qualitative results comparing different loss functions on some challenging cases. Although all models have the capacity to differentiate leaf node classes, for simplicity, we visualise predictions at the top-level nodes using the hierarchy in \figref{fig:hierarchy}. 
We present results at $\tau_0 = 0$ to illustrate the outcome without OOD segmentation for the cross-entropy baseline. 
For the Wasserstein-based loss, we include results using ground distance matrices $M_{t}$ and $M_{h}$ for comparison.
Comparing against $M_{\ell}$ baseline, $\loss_{\WASSCE}$ and $\loss_{\TCE}$ show qualitative results that are more semantically plausible in terms of differentiating normal and abnormal tissues. 
For other classes such as vascular ones, they also show improved segmentation performance by reducing false positive prediction.

\section{Conclusion}
We propose two semantically driven loss functions for sparsely supervised segmentation, relying on a tree-structured label space defined by domain experts.
Both Wasserstein+CE and tree-based semantic CE losses leverage prior knowledge of inter-class relationships:
the former represents these relationships through a distance matrix in label space, while the latter extends the standard cross-entropy loss to incorporate weighted probabilities aggregated at each node in the tree.
Additionally, we integrate these loss functions into an OOD segmentation framework, thereby enabling pixel-level OOD detection without compromising performance on ID data.

Regarding the optimal weighting of hierarchical levels, our experiments on four distance matrices reveal that top-level weights exert the greatest influence on performance when the evaluation is conducted at the corresponding level.
Furthermore, we found that the hierarchical weighting scheme further enhances performance, achieving state-of-the-art results on the dataset for both top-level and leaf node labels.
Moreover, error analysis and qualitative evaluations demonstrate that these approaches offer improved differentiation between normal and abnormal tissues compared with standard cross-entropy baselines.

\begin{credits}
\subsubsection{\discintname}
TV and JS are co-founders and shareholders of Hypervision Surgical Ltd, London, UK.
The authors have no other relevant interests to declare. 
This project received funding by the National Institute for Health and Care Research (NIHR) under its Invention for Innovation (i4i) Programme [NIHR202114].
The views expressed are those of the author(s) and not necessarily those of the NIHR or the Department of Health and Social Care.
This work was supported by core funding from the Wellcome/EPSRC [WT203148/Z/16/Z; NS/A000049/1].
OM is funded by the EPSRC DTP [EP/T517963/1].
\end{credits}

%
%
%

\printbibliography

\end{document}